\documentclass[10pt,twocolumn,letterpaper]{article}

\usepackage[pagenumbers]{cvpr} 

\usepackage{booktabs}
\usepackage{multirow}
\usepackage{graphicx}
\usepackage{threeparttable}
\usepackage{colortbl}
\usepackage{pifont}
\usepackage{adjustbox}

\newcommand{\rendered}{\cellcolor{black!10}}
\newcommand{\notranked}[1]{{\textcolor[rgb]{0.6,0.6,0.6}{#1}}}
\definecolor{agreeA}{HTML}{F4CCCC}
\definecolor{agreeB}{HTML}{FCE5CD}
\definecolor{agreeC}{HTML}{FFF2CC}
\definecolor{agreeD}{HTML}{D9EAD3}

\definecolor{agreeTA}{HTML}{C0392B}
\definecolor{agreeTB}{HTML}{D35400}
\definecolor{agreeTC}{HTML}{B7950B}
\definecolor{agreeTD}{HTML}{1E8449}
\newcommand{\agreetext}[2]{{\footnotesize\color{#1}\text{#2}}}
\newcommand{\renderedtext}[1]{{\setlength{\fboxsep}{1pt}\colorbox{black!10}{\strut #1}}}

\usepackage{etoolbox}
\patchcmd{\thebibliography}
  {\settowidth}
  {\setlength{\itemsep}{0pt plus 0.3pt}\setlength{\parsep}{0pt}\settowidth}
  {}{}

\usepackage{tikz}
\usetikzlibrary{arrows.meta}

\makeatletter
\newcommand{\mypara}{%
  \@startsection{paragraph}{4}%
  {\z@}{0.4ex \@plus .4ex \@minus .4ex}{-1em}%
  {\normalfont\normalsize\bfseries}%
}
\makeatother

\tikzset{pipearrow/.style={line width=0.4pt, line cap=round, >={To[length=2.2pt, width=2.4pt]}}}
\DeclareRobustCommand{\forkto}{\mathrel{\tikz[baseline=-0.55ex, pipearrow]{
  \draw (0,0) -- (0.35em,0);
  \draw[->] (0.35em,0) -- (1.3em,0.38em); \draw[->] (0.35em,0) -- (1.3em,0); \draw[->] (0.35em,0) -- (1.3em,-0.38em);}}}

\newsavebox{\variantpipe}\newsavebox{\variantlineii}\newsavebox{\variantname}
\newcommand{\variant}[2]{%
\sbox{\variantpipe}{#1}\sbox{\variantname}{#2}%
\begin{tabular}[c]{@{}c@{}}\usebox{\variantpipe}\\\usebox{\variantname}\end{tabular}}
\newcommand{\variantthree}[3]{%
\sbox{\variantpipe}{#1}\sbox{\variantlineii}{#2}\sbox{\variantname}{#3}%
\begingroup\dimen0=\wd\variantpipe
\ifdim\wd\variantlineii>\dimen0 \dimen0=\wd\variantlineii \fi
\ifdim\wd\variantname>\dimen0 \dimen0=\wd\variantname \fi
\begin{tabular}[c]{@{}l@{}}%
\makebox[\dimen0][c]{\usebox{\variantpipe}}\\%
\makebox[\dimen0][c]{\usebox{\variantlineii}}\\%
\makebox[\dimen0][c]{\usebox{\variantname}}%
\end{tabular}\endgroup}

\newcommand{\synset}{\textsc{SynTest}}
\newcommand{\wildset}{\textsc{WildTest}}

\newcommand{\vI}{\ensuremath{\mathbf{I}}}  
\newcommand{\vC}{\ensuremath{\mathbf{C}}}  
\newcommand{\vM}{\ensuremath{\mathbf{M}}}  
\newcommand{\vR}{\ensuremath{\mathbf{R}}}  
\newcommand{\vD}{\ensuremath{\mathbf{D}}}  
\newcommand{\vS}{\ensuremath{\mathbf{S}}}  
\newcommand{\vX}{\ensuremath{\mathbf{X}}}  
\newcommand{\vY}{\ensuremath{\mathbf{Y}}}  
\newcommand{\impM}[2][]{\boldsymbol{\mu}_{#2}^{#1}}

\makeatletter
\def\abstract{%
  \iftoggle{cvprpagenumbers}{}{\thispagestyle{empty}}%
  \centerline{\large\bf Abstract}%
  \vspace*{9pt}\noindent\it\ignorespaces}
\makeatother



\AtBeginDocument{%
  \setlength{\abovedisplayskip}{4pt plus 2pt minus 2pt}%
  \setlength{\belowdisplayskip}{4pt plus 2pt minus 2pt}%
  \setlength{\abovedisplayshortskip}{4pt plus 2pt minus 2pt}%
  \setlength{\belowdisplayshortskip}{4pt plus 2pt minus 2pt}%
}

\definecolor{cvprblue}{rgb}{0.21,0.49,0.74}
\usepackage[pagebackref,breaklinks,colorlinks,allcolors=cvprblue]{hyperref}

\def\confName{CVPR}
\def\confYear{2027}

\title{ReMeDi: Single-Image PBR Material Estimation via \\Reparameterization between Metal and Dielectric Hypotheses}

\author{
Fan Fei\textsuperscript{1,2,3}~~~~~Xin Huang\textsuperscript{3}~~~~~Yixin Zhu\textsuperscript{3,4}~~~~~Jiawei Lu\textsuperscript{3}~~~~~Xin Yang\textsuperscript{3}\\
Jinglin Liang\textsuperscript{3}~~~~~Ruining Tang\textsuperscript{3}~~~~~Yifei Feng\textsuperscript{3}~~~~~~Boxin Shi\textsuperscript{1,2\#}~~~~~Chunchao Guo\textsuperscript{3\#}
\\\vspace{-3pt} \normalsize \textsuperscript{1}State Key Laboratory of Multimedia Information Processing, School of Computer Science, Peking University 
\\\vspace{-3pt} \normalsize \textsuperscript{2}National Engineering Research Center of Visual Technology, School of Computer Science, Peking University 
\\\vspace{-3pt} \normalsize 
\textsuperscript{3}Tencent  \quad 
\textsuperscript{4}Nanjing University \quad
\vspace{-7pt}
}

\begin{document}

\twocolumn[{%
    \renewcommand\twocolumn[1][]{#1}%
    \maketitle
    \begin{center}
        \centering
        \captionsetup{type=figure}
        \vspace{-1em}
        \includegraphics[width=\linewidth]{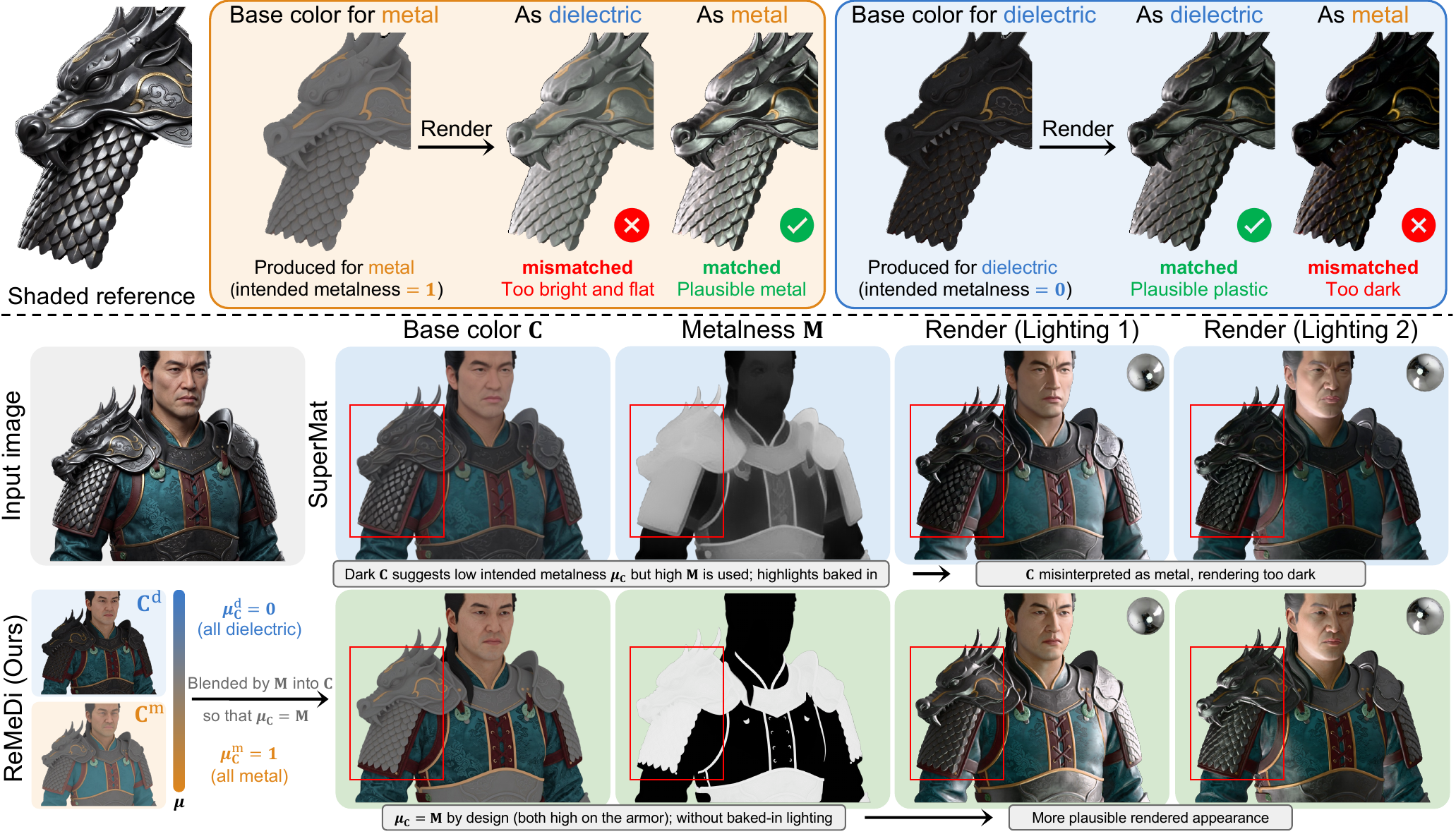}
        \vspace{-1.7em}
        \caption{\textbf{Top:} Any map that determines surface reflectance, such as base color $\vC$, is produced with a decision on the metalness of the material it means to describe, which we call its ``intended metalness'' $\impM{}$.
        Rendering it with a metalness map $\vM$ that does not match $\impM{\vC}$ misinterprets the material.
        \textbf{Middle:} Methods that produce $\vC$ directly may cause such a mismatch, \eg, SuperMat~\cite{hong2025supermat} combines a dark base color meant for a dielectric with high metalness, which renders overly dark.
        \textbf{Bottom:} ReMeDi reparameterizes $\vC$ as an $\vM$-weighted blend of its variants $\vC^{\mathrm{d}}$ and $\vC^{\mathrm{m}}$ under the dielectric and the metal hypothesis, estimated by separate networks, so that $\impM{\vC} = \vM$ by design.
        It also estimates each map more faithfully, \eg, without highlights baked into $\vC$.
        }
        \label{fig: teaser}
    \end{center}
}]

{\renewcommand{\thefootnote}{}%
\footnotetext{This work was done when Fan Fei and Yixin Zhu were interning at Tencent Hunyuan3D.}}

{\renewcommand{\thefootnote}{$\dagger$}%
\footnotetext{Corresponding authors: {\tt\small chunchaoguo@tencent.com, shiboxin@pku.edu.cn}}}

\begin{abstract}
Physically based rendering (PBR) material estimation facilitates relighting, material editing, and 3D content creation.
As the reflectance of a surface heavily depends on whether it is a metal or a dielectric, producing any map that determines albedo involves a decision on metalness, which we call the ``intended metalness'' of the map.
Rendering maps with mismatched intended metalness yields physically implausible or misinterpreted materials.
The metallic--roughness workflow makes metalness explicit and thus encourages base color to match it, but does not enforce this match, especially for automated methods directly predicting each map.
Therefore, we propose ReMeDi, which enforces the match through two components.
First, we reparameterize base color as a metalness-weighted blend of its two variants, estimated under the all-metal and all-dielectric hypothesis, respectively.
Second, as available supervision cannot teach a single network both hypotheses at every pixel, we restrict each of two separate networks to one hypothesis, so that each variant is ensured to follow its hypothesis.
Besides enforcing the match, this also improves each estimated material map.
Experiments show that ReMeDi outperforms state-of-the-art methods in both material accuracy and relighting fidelity.
Project page: \url{https://costrice.github.io/remedi/}
\end{abstract}    
\vspace{-1em}
\section{Introduction}
\label{sec: intro}

Physically based rendering (PBR) materials, represented as spatially varying bidirectional reflectance distribution function (SVBRDF) maps, enable image-space relighting \cite{zeng2024rgbx, liang2025diffusionrenderer, li2025idarb} and material editing \cite{li2020inverse, wang2026materialist}, provide tileable textures \cite{zhou2022tilegen, vecchio2024matfuse, ying2025chord}, and paint 3D assets with view- and lighting-dependent appearance \cite{huang2025material, wu2026matmart, langsteiner2026matspray, wang2024boosting, zhu2026muma, fei2025pacture}.
Recovering them from a single image, a readily available input whether photographed or generated, has therefore long been an important task \cite{aittala2016reflectance, li2017modeling, li2018learning, deschaintre2018single, kocsis2024intrinsic, hong2025supermat, huang2025dualmat}.

The reflective property of a surface heavily depends on whether it is a metal or a dielectric (\ie, its metalness):
A dielectric reflects light diffusely in its own color and specularly with a weak, colorless reflectance of about 4\%, whereas a metal has no diffuse reflection and a strong, often colored specular reflectance \cite{lagarde2014moving}.
Therefore, any map that determines albedo, whether authored or estimated and in whichever SVBRDF parameterization, is produced with a decision on metalness.
For each such map $\vX$, we call this often implicit decision its ``intended metalness'' $\impM{\vX}$, \ie, the metalness of the material that the map's producer means to describe with the map.
For example, a diffuse albedo map $\vD$ produced to describe a painted surface has $\impM{\vD} = \mathbf{0}$.

We note that the intended metalness of different maps describing the same surface may not match, and \textbf{this mismatch may yield physically implausible or misinterpreted materials.}
The specular--glossiness (SG) workflow stores diffuse albedo $\vD$ and specular albedo $\vS$ directly, leaving $\impM{\vD}$ and $\impM{\vS}$ implicit and unconstrained by each other (\cref{tab: intended metalness}, SG).
As a result, it may describe a material that is highly dielectric and metallic at once, \eg, a bright $\vD$ with a colored $\vS$, where the low $\impM{\vD}$ does not match the high $\impM{\vS}$ \cite{mcdermott2018pbr}.
The metallic--roughness (MR) workflow \cite{burley2012physically} makes the decision explicit as a metalness map $\vM$, which is usually binary and directly shows its own intended metalness, \ie, $\impM{\vM} = \vM$ (\cref{tab: intended metalness}, MR direct).
The renderer then interprets a base color map $\vC$ as diffuse albedo where $\vM = \mathbf{0}$ and as specular albedo where $\vM = \mathbf{1}$.
This parameterization encourages $\impM{\vC}$ to match $\vM$, as authors are required to produce $\vM$ and can directly refer to it when producing $\vC$.
Nevertheless, the match is not enforced.
The surface can look darker than expected if a $\vC$ intended as diffuse albedo ($\impM{\vC} = \mathbf{0}$) is rendered as metal ($\vM = \mathbf{1}$), and brighter in the opposite case (\cref{fig: teaser}, top).
The mismatch problem is more pronounced for learning-based methods that directly produce each map~\cite{deschaintre2018single, zhou2022tilegen, ying2025chord, zhu2026muma, zeng2024rgbx, huang2025material} than for human artists, as nothing but the learned data distribution, which only loosely reflects the match, relates the intended metalness of different maps (\cref{fig: teaser}, middle).

\begin{table}[t]
\centering
\caption{
Intended metalness $\boldsymbol{\mu}$ of the maps produced in the specular--glossiness (SG) workflow, which stores diffuse albedo $\vD$ and specular albedo $\vS$, and the metallic--roughness (MR) workflow, which stores $\vM$ and $\vC$.
Existing MR methods produce $\vC$ directly, whereas ReMeDi (1) reparameterizes it by its two variants and (2) estimates them with separate hypothesis-specific networks.
\renderedtext{Shaded} entries are the final maps used by the renderer, ``?'' denotes intended metalness with uncertainty, and ``Match'' indicates how the methods make the $\boldsymbol{\mu}$ of the maps match.
}
\vspace{-0.5em}
\label{tab: intended metalness}
\small
\setlength{\tabcolsep}{4pt}
\begin{tabular}{@{}l ccccc c@{}}
\toprule
\multirow{2.3}{*}{Production} & \multicolumn{5}{c}{$\boldsymbol{\mu}$ of the produced maps} & \multirow{2.3}{*}{Match of $\boldsymbol{\mu}$} \\
\cmidrule(lr){2-6}
& $\vD$ & $\vS$ & $\vC^{\mathrm{d}}$\,/\,$\vC^{\mathrm{m}}$ & $\vM$ & $\vC$ & \\
\midrule
SG & \rendered ? & \rendered ? & -- & -- & -- & \agreetext{agreeTA}{Unconstrained} \\
MR (direct) & -- & -- & -- & \rendered $\vM$ & \rendered ? & \agreetext{agreeTB}{Encouraged} \\
MR + (1) & -- & -- & $\mathbf{0}$?\,/\,$\mathbf{1}$? & \rendered $\vM$ & \rendered $\vM$? & \agreetext{agreeTC}{Conditional} \\
MR + (1)(2) & -- & -- & $\mathbf{0}$\,/\,$\mathbf{1}$ & \rendered $\vM$ & \rendered $\vM$ & \agreetext{agreeTD}{Enforced} \\
\bottomrule
\end{tabular}
\end{table}

To this end, we propose ReMeDi to estimate material maps from a single image via \underline{Re}parameterization between \underline{Me}tal and \underline{Di}electric hypotheses, which \textbf{enforces the match by design} through two components, (1) a reparameterization of base color between all-metal and all-dielectric hypotheses and (2) a hypothesis-specific supervision and estimation pipeline (\cref{tab: intended metalness}, MR + (1)(2) and \cref{fig: teaser}, bottom).

As component (1), instead of producing $\vC$ directly with its $\impM{\vC}$ left implicitly to the network, we reparameterize $\vC$ as a blend of its two variants $\vC^{\mathrm{m}}$ and $\vC^{\mathrm{d}}$, meant to be produced under the all-metal hypotheses ($\impM[\mathrm{m}]{\vC} = \mathbf{1}$) and the all-dielectric ($\impM[\mathrm{d}]{\vC} = \mathbf{0}$), respectively, according to $\vM$.
We form $\vC$ by selecting $\vC^{\mathrm{m}}$ where $\vM = \mathbf{1}$ and $\vC^{\mathrm{d}}$ where $\vM = \mathbf{0}$, and generalize this selection to intermediate values of $\vM$, interpreting these values as metal coverage.
Through this reparameterization, as long as $\impM[\mathrm{m}]{\vC} = \mathbf{1}$ and $\impM[\mathrm{d}]{\vC} = \mathbf{0}$ are actually fulfilled, $\impM{\vC}$ is ensured to match $\vM$.

Component (1) alone, however, is not sufficient, as ensuring the premise instead of leaving it to chance is not straightforward (\cref{tab: intended metalness}, MR + (1)).
As no existing asset provides two sets of material maps that have opposite metalness and render to the same appearance, each hypothesis can only be supervised on pixels where it holds, leaving the predictions elsewhere unconstrained.
Thus, a single network that learns to estimate under both hypotheses is not guaranteed to fully follow the hypothesis required by $\vC^{\mathrm{m}}$ or $\vC^{\mathrm{d}}$.
As component (2), we guarantee this by restricting each of two separate networks to learn under a single and fixed hypothesis, so that it can only estimate under that hypothesis, whether it holds at a pixel or not.
This ensures that the premise is fulfilled, \ie, $\impM[\mathrm{m}]{\vC} = \mathbf{1}$ and $\impM[\mathrm{d}]{\vC} = \mathbf{0}$.

As shown in the bottom of \cref{fig: teaser}, the reparameterization and the hypothesis-specific estimation together enforce the intended metalness of the blended $\vC$ to be $\vM$ by design, eliminating the mismatch and its consequences.
Moreover, the separate networks also improve the estimated materials maps, mitigating problems such as highlights baked into base color.
This is because metals and dielectrics follow different distributions of base color and roughness, thus calling for different ways of inferring the intrinsic material from the observed appearance.
Extensive experiments demonstrate that ReMeDi outperforms existing methods in both material accuracy and relighting fidelity, reducing the base color and relighting errors by 2.86\,dB and 2.74\,dB, respectively, and validate the necessity of both components.
\section{Related Work}
\label{sec: related}

As shown in \cref{tab: related work}, in this section, we review SVBRDF parameterizations and summarize methods producing them, mostly from a single image.

\mypara{SVBRDF parameterizations.}
SVBRDF maps typically parameterize a microfacet BRDF~\cite{cook1982reflectance, walter2007microfacet}, which sums a diffuse lobe and a specular lobe shaped by the normal and roughness ($\vR$) at each surface point.
The SG workflow~\cite{mcdermott2018pbr} specifies the reflectance of the two lobes with two independent color maps $\vD$ and $\vS$, which do not prevent physically implausible combinations.
The MR workflow~\cite{burley2012physically, karis2013real} instead derives both from $\vC$ and $\vM$, using $\vM$ as the weight for linearly blending dielectric and metallic reflection models.
It builds in that dielectrics share an achromatic $\vS$ of about 0.04 and metals have no diffuse reflection, leading to
\begin{equation}
\vD = \mathrm{lerp}(\vC, \mathbf{0}; \vM), \quad \vS = \mathrm{lerp}(\mathbf{0.04}, \vC; \vM), \label{eq: cm to ds}
\end{equation}
where $\mathrm{lerp}(\cdot, \cdot; \cdot)$ is linear interpolation.
This makes plausible materials easier to author and the maps more compact, as a single-channel $\vM$ replaces the three-channel $\vS$~\cite{mcdermott2018pbr}.
MR has thus become the default in asset formats such as glTF~\cite{gltf2spec}, USD~\cite{usdpreviewsurface}, and OpenPBR~\cite{openpbr}.

\mypara{SVBRDF estimation under SG.}
Methods in the SG workflow (\cref{tab: related work}, rows~1--4) estimate $\vR$ together with $\vD$, $\vS$, or both from a single image.
Most predict $\vD$ and $\vS$ jointly, either pixel-aligned with the input~\cite{deschaintre2018single, guo2021highlight, zhou2021adversarial, zhou2022look, vecchio2021surfacenet, nie2025auxiliary, xing2025diffusion} or as tileable textures~\cite{zhou2022tilegen, vecchio2024matfuse}, while others predict only $\vD$ and fix $\vS$ to the dielectric value of 0.04~\cite{li2018learning, chen2024intrinsicanything, wei2025sir, lopes2024material}, which cannot represent metals.
Neither the workflow nor the pipelines constrain $\impM{\vD}$ and $\impM{\vS}$ to match.

\mypara{SVBRDF estimation under MR.}
Unlike $\vD$ and $\vS$, $\vC$ takes its meaning from $\vM$,  which encourages, but not enforces, $\impM{\vC}$ to match $\impM{\vM} \equiv \vM$.
We thus organize methods in the MR workflow (rows~5--12), which estimate $\vR$ together with $\vC$ and $\vM$, by when $\vM$ is determined relative to $\vC$, \ie, how they relate the two.
Several first produce $\vC$ and add $\vM$ afterwards (rows~5--7).
CHORD~\cite{ying2025chord} motivates this order by noting that $\vC$ most resembles the input among all attributes, and searches over $\vM$ of 0 and 1 per pixel while keeping the same $\vC$.
MuMA~\cite{zhu2026muma} generates $\vC$ along with a shaded image, from which IDArb~\cite{li2025idarb} derives $\vM$, and Boosting3D~\cite{wang2024boosting} assigns $\vM$ to segmented parts with values suggested by a vision-language model.
Many others produce $\vC$ and $\vM$ jointly (rows~8--10), whether pixel-aligned~\cite{zeng2024rgbx, zeng2026rgbxnext, kocsis2024intrinsic, hong2025supermat, huang2025dualmat, li2025idarb, ke2025marigold, liang2025diffusionrenderer}, tileable~\cite{vecchio2024controlmat, vecchio2026stablematerials, luo2026matpedia}, or as UV textures~\cite{huang2025material, fei2025pacture, he2025materialmvp, chen2025primdiff, zhu2024mcmat}.
A few encourage $\impM{\vC}$ to match $\vM$ by first predicting $\vM$ then conditioning $\vC$ on $\vM$ (rows~11--12). 
Rhee and Lee~\cite{rhee2022estimating} route the whole image to a dedicated metal or non-metal network, which is the closest to our method that restricts a network to a single kind of material, but their decision is per image.
In contrast, ReMeDi applies the two hypotheses per pixel by $\vM$ and thus enforces the match by design, even on images containing both kinds of materials.

\begin{table}[t]
\centering
\caption{
Representative methods that produce SVBRDF maps, mostly from a single image.
The SVBRDF parameterization (leftmost column) is either SG or MR.
The type of the produced SVBRDF map includes maps pixel-wise aligned to the inputs, tileable textures for arbitrary surfaces, and UV maps specific to a given mesh.
In each pipeline, $\to$ feeds the input image $\vI$, or the output of a previous network, into the next network; $+$ joins maps produced by the same network; and $\Rightarrow$ denotes an analytic step without learned parameters, combining the two hypotheses by metalness (\cref{eq: generalized combination,eq: combination roughness}).
$\forkto$ and the vertical bars $|$ separate the outputs of the three networks of ReMeDi.
}
\begin{threeparttable}
\setlength{\tabcolsep}{3pt}
\small
\resizebox{\columnwidth}{!}{
\begin{tabular}{@{\hspace{2pt}}l@{\hspace{2pt}}r@{\hspace{3pt}}l c l@{}}
\toprule
 & & Representative works & Output type & Pipeline (from $\vI$) \\
\midrule
\multirow{5}{*}[0pt]{\rotatebox[origin=c]{90}{SG workflow}} & (1) & Li \etal~\cite{li2018learning}; \cite{chen2024intrinsicanything, wei2025sir} & Aligned & $\to \vD$ \\[0.8pt]
 & (2) & Material Palette~\cite{lopes2024material} & Tileable & $\to \vD$ \\[0.8pt]
\cmidrule(l){2-5}
 & (3) & Deschaintre \etal~\cite{deschaintre2018single}; & \multirow{2}{*}{Aligned} & \multirow{2}{*}{$\to \vD + \vS$} \\[0.8pt]
 & & \cite{guo2021highlight, zhou2021adversarial, zhou2022look, vecchio2021surfacenet, nie2025auxiliary, xing2025diffusion} & & \\[0.8pt]
 & (4) & TileGen~\cite{zhou2022tilegen}; \cite{vecchio2024matfuse} & Tileable & $\to \vD + \vS$ \\[0.6pt]
\midrule
\multirow{11}{*}{\rotatebox[origin=c]{90}{MR workflow}} & (5) & CHORD~\cite{ying2025chord} & Tileable & $\to \vC \to \vM$ \\
 & (6) & MuMA~\cite{zhu2026muma} & UV map & $\to \vC \to \vM$ \\
 & (7) & Boosting3D~\cite{wang2024boosting} & UV map & $\to \vC \to \vM^{\ast}$ \\
\cmidrule(l){2-5}
 & (8) & RGB$\leftrightarrow$X~\cite{zeng2024rgbx}; \cite{zeng2026rgbxnext, kocsis2024intrinsic} & \multirow{2}{*}{Aligned} & \multirow{2}{*}{$\to \vC + \vM$} \\
 & & \cite{hong2025supermat, huang2025dualmat, li2025idarb, ke2025marigold, liang2025diffusionrenderer} & & \\
 & (9) & ControlMat~\cite{vecchio2024controlmat}; \cite{vecchio2026stablematerials, luo2026matpedia} & Tileable & $\to \vC + \vM$ \\
 & (10) & Material Anything~\cite{huang2025material}; & \multirow{2}{*}{UV map} & \multirow{2}{*}{$\to \vC + \vM$} \\
 & & \cite{fei2025pacture, he2025materialmvp, chen2025primdiff, zhu2024mcmat} & & \\
\cmidrule(l){2-5}
 & (11) & PBR3DGen~\cite{wei2026pbr3dgen} & UV map & $\to \vM^{\ast} \to \vC + \vM$ \\
 & (12) & Rhee and Lee~\cite{rhee2022estimating} & Aligned & $\to \vM^{\ast} \to \vC$ \\
\cmidrule(l){2-5}
 & (13) & \textbf{ReMeDi (Ours)} & Aligned & $\forkto \vC^{\mathrm{d}} \,|\, \vM \,|\, \vC^{\mathrm{m}} \Rightarrow \vC$ \\
\bottomrule
\end{tabular}
}
\begin{tablenotes}\footnotesize
\item[*] decided per image or per part rather than per pixel.
\end{tablenotes}
\end{threeparttable}
\label{tab: related work}
\end{table}
\section{Proposed Method}
\label{sec: method}

\begin{figure*}[t]
\centering
\includegraphics[width=\linewidth]{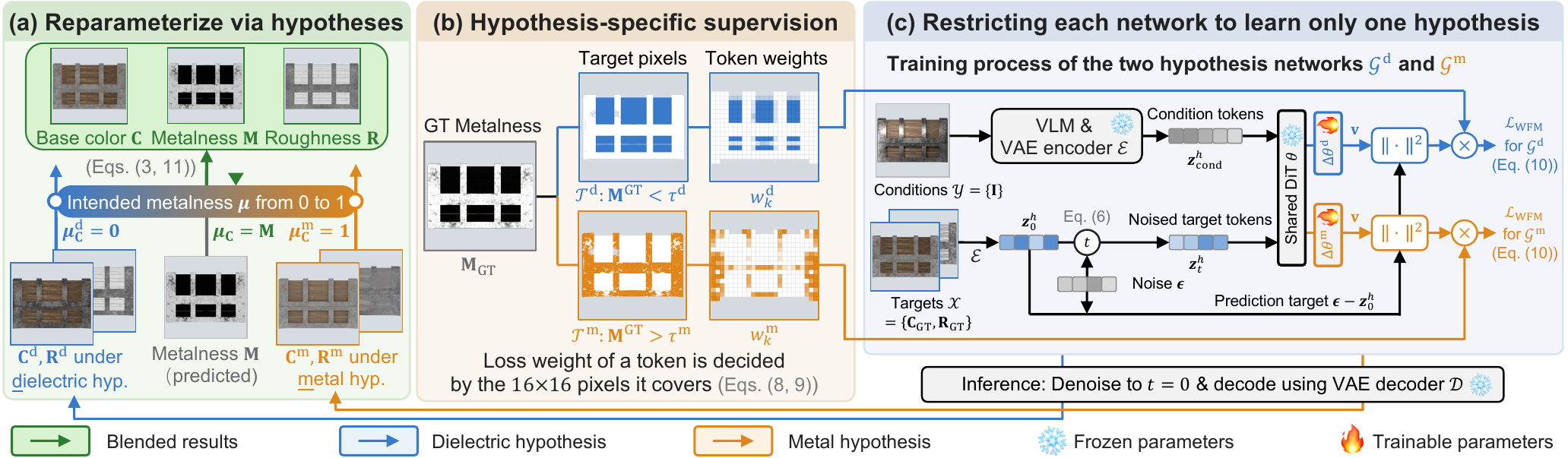}
\caption{
Overview of ReMeDi.
(a) ReMeDi reparameterizes base color $\vC$ as a blend of its variants under the dielectric and metal hypotheses, weighted by metalness $\vM$, so that the intended metalness of $\vC$ matches $\vM$ by construction (\cref{subsec: reparam}). Roughness $\vR$ is blended in the same way.
(b) Each variant is supervised only where its hypothesis holds. 
Ground-truth (GT) metalness gives the target pixels $\mathcal{T}^{h}$ ($h\in \{\mathrm{d},\mathrm{m}\}$), whose fraction in each token determines the token weight $w_k^h$ (\cref{subsec: supervision}).
(c) As the supervision alone does not guarantee that the estimated variants indeed follow their hypothesis on every pixel, we use separate networks, each with its own low-rank adaptation (LoRA) weights $\Delta\theta$, and restrict each network to learn only one hypothesis (\cref{subsec: separate network}). 
The metalness predictor is also trained with LoRA as an independent network, similar to the hypothesis networks but with target maps set to $\{\vM_{\rm GT}\}$ and token weights set to $1$.
}
\label{fig: method}
\end{figure*}

Given a single image $\vI \in [0, 1]^{H \times W \times 3}$ of an object, ReMeDi estimates pixel-aligned base color $\vC \in [0, 1]^{H \times W \times 3}$, metalness $\vM \in [0, 1]^{H \times W}$, and roughness $\vR \in [0, 1]^{H \times W}$ in the MR workflow (\cref{fig: method}).
We first show how to reparameterize $\vC$ under two fixed hypotheses so that its intended metalness $\impM{\vC}$ matches the used metalness map $\vM$ (component (1), \cref{subsec: reparam}).
After introducing the conditional generation model on which our networks are built (\cref{subsec: preliminary}), we describe how each variant is supervised only where its hypothesis holds (\cref{subsec: supervision}).
We then show how separate networks ensure that each variant follows its hypothesis at every pixel (component (2), \cref{subsec: separate network}), followed by the implementation details (\cref{sec: method implementation}).

\subsection{Reparameterization between Hypotheses}
\label{subsec: reparam}

As discussed in previous sections, base color is the diffuse albedo of a dielectric but the specular albedo of a metal, so a base color map $\vC$ comes with its intended metalness map $\impM{\vC}$.
When base color $\vC$ is produced directly, its $\impM{\vC}$ is left to the producer and may not match the explicit metalness map $\vM$, leading to materials that appear too dark or too bright.
To eliminate this potential mismatch, we reparameterize $\vC$ as a combination of two variants by $\vM$ so that $\impM{\vC} = \vM$ holds as long as a premise is fulfilled.

\mypara{Picking one from two hypotheses about metalness.}
We first introduce two variants of base color, $\vC^{\mathrm{d}}$ and $\vC^{\mathrm{m}}$, which describe the input as if every pixel were dielectric or metal, respectively.
Thus, their intended metalness is $\mathbf{0}$ and $\mathbf{1}$, respectively, whatever the actual material is.
$\vC$ then picks at every pixel $p$ the exact hypothesis that $\vM$ indicates:
\begin{equation}
\vC(p) = \begin{cases}
    \vC^{\mathrm{d}}(p), & \text{where } \vM(p) = 0 \\
    \vC^{\mathrm{m}}(p), & \text{where } \vM(p) = 1
\end{cases}.
\label{eq: binary combination}
\end{equation}
This ensures $\impM{\vC} = \vM$ where $\vM$ is binary.
Note that $\vC^{\mathrm{d}}$ and $\vC^{\mathrm{m}}$ do not equal the diffuse albedo $\vD$ or specular albedo $\vS$, which depend on the actual metalness. 
For example, $\vD$ should be $0$ at a metal pixel while $\vC^{\mathrm{d}}$ is meant to describe this pixel as dielectric and predict its diffuse albedo under that hypothesis, which is generally non-zero.

\mypara{Generalization to intermediate metalness.}
When $\vM$ is not binary, combining the two variants of $\vC$ so that $\impM{\vC}$ is still $\vM$ is not as trivial as in \cref{eq: binary combination}.
In the MR workflow \cite{burley2012physically,karis2013real}, intermediate metalness values indicate a spatial mixture of dielectric and metal surfaces, such as material boundaries and partially rusted metals.
The value of metalness is interpreted as the fraction covered by the metal, and thus serves as the linear blending weight between the two reflection models in rendering (\cref{eq: cm to ds}).
The same linear blending of different materials is also used in material-layering works~\cite{drehwald2023oneshot, unity_layered_lit}.
Thus, we follow this convention and use $\vM$ to linearly blend the two base color variants at every pixel, which generalizes \cref{eq: binary combination}:
\begin{equation}
\vC(p) = \mathrm{lerp}(\vC^{\mathrm{d}}(p), \vC^{\mathrm{m}}(p); \vM(p)), \quad \forall p,
\label{eq: generalized combination}
\end{equation}
The blended $\vC$ thus has $\impM{\vC} = \vM$, as long as $\vC^{\mathrm{d}}$ and $\vC^{\mathrm{m}}$ indeed follow their respective hypotheses, which we enforce by the pipeline design described in \cref{subsec: supervision,subsec: separate network}.

\subsection{Preliminary: Conditional Generation}
\label{subsec: preliminary}

Our networks are built on a pretrained diffusion transformer (DiT) for conditional image generation~\cite{qwenimage21}, fine-tuned with low-rank adaptation (LoRA)~\cite{hu2022lora}.
In this subsection, we describe a single module $\mathcal{G}^{h}$ that generates a set of target maps $\mathcal{X}^{h}$ from a set of condition images $\mathcal{Y}^{h}$.
For all modules of ReMeDi, $\mathcal{Y}^{h} = \{\vI\}$, while some variants in \cref{sec: exp ablation} additionally condition on $\vM$.

\mypara{Image tokens.}
The DiT generates images as sequences of tokens, which a frozen variational autoencoder (VAE)~\cite{kingma2014auto} encodes from images and decodes back.
Its encoder $\mathcal{E}$ downsamples an image $\vX$ by a factor of 16 into $d = 64$ channels, so that $\mathcal{E}(\vX) \in \mathbb{R}^{\frac{H}{16} \times \frac{W}{16} \times d}$, \ie, $N = HW / 16^2$ tokens, each covering a $16 \times 16$ pixel patch.
We replicate single-channel maps such as roughness and metalness to three channels as we empirically find that keeping them as separate maps instead of a combined map gives sharper and cleaner generation results.

\mypara{Conditioning.}
The DiT processes text and image tokens in a single sequence, whose prefix holds the condition tokens.
A frozen vision-language model (VLM) encoder~\cite{bai2025qwen3vl} jointly encodes $\vI$ and a fixed text prompt into a sequence of embeddings, of which those at the text positions, $\mathbf{e}_\text{text}$, are kept, and those at the image positions are replaced by $\mathcal{E}(\vI)$.
Any other condition image is encoded by the VAE into a further block, so that the condition tokens of module $h$ are
\begin{equation}
\mathbf{z}_{\mathrm{cond}}^{h} = \mathrm{concat}\big\{\, \mathbf{e}_\text{text},\ \mathcal{E}(\vY) : \vY \in \mathcal{Y}^{h} \,\big\}.
\label{eq: condition}
\end{equation}

\mypara{Target.}
The original backbone generates only a single target image after the prefix.
We extend it to multiple target maps by appending their tokens after the prefix, each map with its own positional embedding, and letting all of them attend to the whole prefix and to each other, so that the maps of a module are generated jointly.
Specifically, the target tokens $\mathbf{z}_0^{h}$ are formed by encoding each target map into a grid of tokens and concatenating the flattened grids:
\begin{equation}
\mathbf{z}_0^{h} = \mathrm{concat}\big\{\, \mathcal{E}(\vX) : \vX \in \mathcal{X}^{h} \,\big\} \in \mathbb{R}^{|\mathcal{X}^{h}| N \times d}.
\label{eq: target tokens}
\end{equation}

\mypara{Denoising.}
Following flow matching~\cite{lipman2023flow}, we sample a time $t \in [0, 1]$ and standard Gaussian noise $\boldsymbol{\epsilon}$ of the same shape as $\mathbf{z}_0^{h}$ and interpolate into noised targets,
\begin{equation}
\mathbf{z}_t^{h} = (1 - t)\, \mathbf{z}_0^{h} + t\, \boldsymbol{\epsilon} \in \mathbb{R}^{|\mathcal{X}^{h}| N \times d},
\label{eq: interpolation}
\end{equation}
which are then appended to $\mathbf{z}_{\mathrm{cond}}^{h}$. 
The transformer $\mathbf{v}_{\theta, \Delta\theta^{h}}$, with frozen backbone weights $\theta$ and the LoRA weights $\Delta\theta^{h}$ of module $h$, predicts the velocity $\boldsymbol{\epsilon} - \mathbf{z}_0^{h}$ of the target.
At inference, we start from pure noise at $t = 1$, integrate the predicted velocity to $t = 0$, and decode each block of the result $\hat{\mathbf{z}}_0^{h}$ with the frozen VAE decoder $\mathcal{D}$:
\begin{equation}
\hat{\vX} = \mathcal{D}\big(\hat{\mathbf{z}}_0^{h}[\vX]\big) \in [0, 1]^{H \times W \times 3}, \quad \vX \in \mathcal{X}^{h},
\label{eq: decoding}
\end{equation}
where $\hat{\mathbf{z}}_0^{h}[\vX] \in \mathbb{R}^{N \times d}$ is the block corresponding to $\vX$.
As the single-channel maps are replicated before encoding, we average the three channels of their decoded output.

\begin{table*}[ht!]
\centering
\caption{
Quantitative comparison on \synset{} with GT materials.
Each metric is computed per image on the foreground and averaged over all images.
``Relighting'' compares the predicted and GT materials rendered by the same single-view-based renderer under identical normals and lighting.
The best and second-best results are in \textbf{bold} and \underline{underlined}.
}
\vspace{-4pt}
\label{tab: comparison}
\small
\setlength{\tabcolsep}{3pt}
\begin{tabular}{@{}l cccc cc c ccc@{}}
\toprule
& \multicolumn{4}{c}{Base color} & \multicolumn{2}{c}{Metalness} & Roughness & \multicolumn{3}{c}{Relighting} \\
\cmidrule(lr){2-5} \cmidrule(lr){6-7} \cmidrule(lr){8-8} \cmidrule(lr){9-11}
Method & PSNR$\uparrow$ & si-PSNR$\uparrow$ & SSIM$\uparrow$ & LPIPS$\downarrow$ & RMSE$\downarrow$ & Acc.$\uparrow$ & RMSE$\downarrow$ & PSNR$\uparrow$ & SSIM$\uparrow$ & LPIPS$\downarrow$ \\
\midrule
RGB$\leftrightarrow$X~\cite{zeng2024rgbx} & 13.16 & 16.85 & 0.881 & 0.120 & 0.599 & 0.388 & 0.258 & 14.56 & 0.897 & 0.091 \\
IID~\cite{kocsis2024intrinsic} & 14.09 & 17.89 & 0.893 & 0.121 & 0.399 & 0.690 & 0.262 & 15.52 & 0.913 & 0.087 \\
DiffusionRenderer~\cite{liang2025diffusionrenderer} & 15.31 & 18.55 & 0.901 & 0.105 & 0.413 & 0.693 & 0.243 & 16.59 & 0.917 & 0.078 \\
Marigold-IID~\cite{ke2025marigold} & 15.55 & 18.95 & 0.905 & 0.121 & 0.504 & 0.450 & 0.222 & 17.00 & 0.925 & 0.080 \\
IDArb~\cite{li2025idarb} & 16.60 & 18.53 & 0.905 & \underline{0.092} & 0.473 & 0.628 & 0.273 & 17.57 & 0.914 & 0.074 \\
SuperMat~\cite{hong2025supermat} & \underline{18.47} & \underline{20.62} & \underline{0.920} & 0.094 & \underline{0.266} & \underline{0.847} & \underline{0.191} & \underline{19.60} & \underline{0.937} & \underline{0.065} \\
\textbf{ReMeDi (Ours)} & \textbf{21.33} & \textbf{23.79} & \textbf{0.956} & \textbf{0.039} & \textbf{0.197} & \textbf{0.913} & \textbf{0.170} & \textbf{22.34} & \textbf{0.965} & \textbf{0.031} \\
\bottomrule
\end{tabular}
\vspace{-4pt}
\end{table*}

\subsection{Hypothesis-Specific Supervision}
\label{subsec: supervision}

The effectiveness of the reparameterization relies on that the two base color variants, $\vC^{\mathrm{d}}$ and $\vC^{\mathrm{m}}$, follow their respective hypotheses, \ie, describing every pixel as the same kind of material.
However, the ground-truth (GT) material maps of a training asset follow its actual material.
The GT base color map is thus a valid target for a variant only on pixels where its hypothesis holds, \eg, metal pixels for $\vC^{\mathrm{m}}$, and misleads it elsewhere.
We therefore supervise each variant only on regions of its own kind of material.

As the generation operates on tokens, we weight each token by the pixels it covers.
We call a pixel dielectric if its GT metalness is below $\tau^\text{d}=0.2$ and metal if it exceeds $\tau^\text{m}=0.8$, which gives the sets of target pixels $\mathcal{T}^{\mathrm{d}}$ and $\mathcal{T}^{\mathrm{m}}$ of the two hypotheses.
For each token $k$ of a variant map under hypothesis $h \in \{\mathrm{d}, \mathrm{m}\}$, covering the $16 \times 16$ pixels $\mathcal{P}_k$, we compute the fraction of target pixels:
\begin{equation}
f_k^h = \frac{|\mathcal{P}_k \cap \mathcal{F} \cap \mathcal{T}^{h}|}{|\mathcal{P}_k \cap \mathcal{F}|} \in [0, 1],
\label{eq: token fraction}
\end{equation}
where $\mathcal{F}$ is the set of foreground pixels.
Tokens dominated by the target pixels are supervised with the weight
\begin{equation}
w_k^h = \mathrm{clamp}\!\left( \frac{f_k^h - \tau^\text{p}}{1 - \tau^\text{p}},\, 0,\, 1 \right),
\label{eq: token weight}
\end{equation}
where $\tau^\text{p} = 0.9$ is the purity threshold of the target material for a token to participate in training.
We apply $w_k^h$ to get a weighted flow matching objective of the module $\mathcal{G}^{h}$ that generates the variant maps under hypothesis $h$:
\begin{equation}
\mathcal{L}_\text{WFM} = \frac{\sum_k w_k^h \left\| \mathbf{v}_{\theta, \Delta\theta^{h}}(\mathbf{z}_t^{h}, t, \mathbf{z}_{\mathrm{cond}}^{h})_k - (\boldsymbol{\epsilon} - \mathbf{z}_0^{h})_k \right\|^2}{\sum_k w_k^h}.
\label{eq: fm}
\end{equation}
For both variants, the target tokens $\mathbf{z}_0^{h}$ correspond to the same GT map, and their difference comes from $w_k^h$.

\subsection{Restricting One Network to One Hypothesis}
\label{subsec: separate network}

The supervision in \cref{subsec: supervision} leaves the output of each variant on pixels of the other kind of material to the extrapolation of the network.
For example, outputting the actual base color for both variants fits the supervision perfectly, though they do not fully follow the hypotheses.
Thus, a single network that learns under both hypotheses may apply either of them to each variant, and the produced two variants are not ensured to follow their own hypotheses, as shown in \cref{sec: exp ablation}.

Imposing more complete supervisions by constraining both variants on the same pixels would require two sets of material maps that have opposite metalness and render to the same appearance, which existing assets do not provide.
An alternative to constrain every pixel is thus required.

\mypara{Separate hypothesis networks.}
We therefore restrict each network to learn under a single hypothesis, using a dielectric module $\mathcal{G}^{\mathrm{d}}$ and a metal module $\mathcal{G}^{\mathrm{m}}$ built on \cref{subsec: preliminary}, together with a metalness module $\mathcal{G}^{\mathrm{M}}$ described below.
The modules share the frozen backbone but have their own LoRA weights.
As each module never learns to estimate under the other hypothesis, it surely follows its own hypothesis at every pixel, fulfilling the requirement of \cref{subsec: reparam}.

Separate modules also improve the estimated materials of both kinds, as dielectrics and metals follow different distributions of material values and call for different ways to infer the material from appearance.
For example, the metal variant requires stronger ability to identify and remove metallic highlights, while the dielectric variant should focus more on retaining detailed textures in the input.

\mypara{Roughness.}
Roughness $\vR$ keeps its meaning for both dielectrics and metals, but its typical values differ significantly between them, \eg, a high roughness is far more common on dielectrics.
A roughness map thus also carries an intended metalness, though more weakly tied to its values than that of base color.
We therefore let $\mathcal{G}^{\mathrm{d}}$ and $\mathcal{G}^{\mathrm{m}}$ estimate roughness under their own hypotheses as well, \ie, $\mathcal{X}^{\mathrm{d}} = \{\vC^{\mathrm{d}}, \vR^{\mathrm{d}}\}$ and $\mathcal{X}^{\mathrm{m}} = \{\vC^{\mathrm{m}}, \vR^{\mathrm{m}}\}$, and combine the two in the same way as base color:
\begin{equation}
\vR(p) = \mathrm{lerp}(\vR^{\mathrm{d}}(p), \vR^{\mathrm{m}}(p); \vM(p)), \quad \forall p.
\label{eq: combination roughness}
\end{equation}

\mypara{Metalness predictor.}
The metalness module $\mathcal{G}^{\mathrm{M}}$, with $\mathcal{X}^{\mathrm{M}} = \{\vM\}$, estimates the metalness that blends the two hypotheses.
Besides \cref{eq: fm}, a binarization penalty keeps its estimate near-binary where the GT is binary:
\begin{equation}
\mathcal{L}_\text{bin} = \frac{1}{|\Omega|} \sum_{p \in \Omega} \left| \hat{\vM}_0(p) - [\vM_\text{GT}(p) > 0.5] \right|,
\label{eq: binarization}
\end{equation}
where $\hat{\vM}_0$ is the metalness decoded from the clean estimate $\mathbf{z}_t^{\mathrm{M}} - t\, \mathbf{v}_{\theta, \Delta\theta^{\mathrm{M}}}$, $\Omega \subseteq \mathcal{F}$ is the set of foreground pixels whose GT metalness is clearly binary, \ie, below $\tau^\text{d}$ or above $\tau^\text{m}$, and $[\cdot]$ is the indicator function.
The penalty is applied only at small $t$, where $\hat{\vM}_0$ is reliable.
As pixels with intermediate GT metalness are left out of $\Omega$, the predicted metalness remains continuous where the GT is.
The metalness module is trained with $\mathcal{L}_\text{WFM} + \lambda\, \mathcal{L}_\text{bin}$, where $w_k^{\mathrm{M}} = 1$ on all foreground tokens and $\lambda$ balances the two terms.

\subsection{Implementation Details}
\label{sec: method implementation}

\mypara{Training and inference.}
We train each module separately at a resolution of $1024 \times 1024$, with LoRA of rank 64 on the transformer, AdamW at a learning rate of $5 \times 10^{-5}$ after a linear warm-up of 500 iterations, and $\lambda = 0.3$.
Each module is trained for 30K iterations with a batch size of 1 per GPU and a sequence parallelism (SP) size of 2, on 24 NVIDIA H20 GPUs.
The training set contains about 95K assets.
At inference, the three modules take the same $1024 \times 1024$ input and use 20 sampling steps each, which takes about 60\,s in total for an average case on a single GPU.
\section{Experiments}
\label{sec: experiments}

\begin{figure*}[t]
\centering
\includegraphics[width=\linewidth]{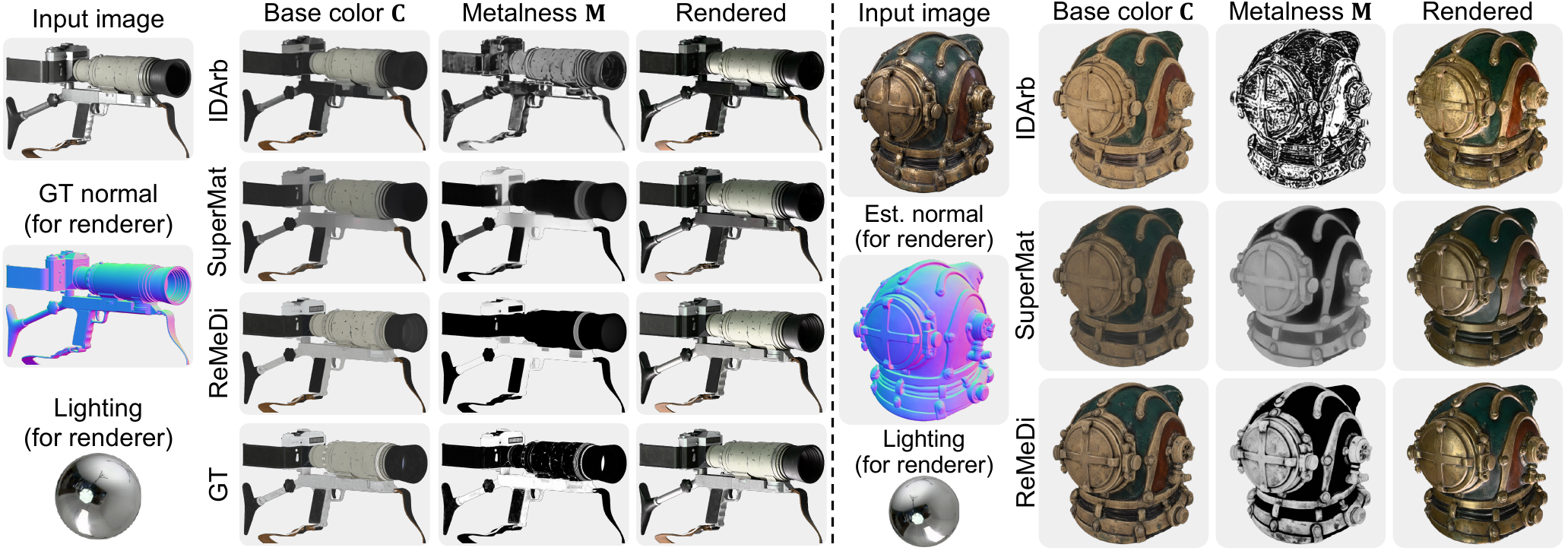}
\caption{
Qualitative comparison on \synset{} (left, with GT) and \wildset{} (right, without GT).
For each example, we show the estimated base color, metalness, and their relit results using GT normals for \synset{} or estimated normals by \cite{kong2026moge3} for \wildset{}.
}
\label{fig: comparison}
\end{figure*}

The experiments are conducted on two test sets: \synset{} containing rendered images of 3D assets with GT material maps and \wildset{} containing in-the-wild images without GT material (\cref{sec: exp setup}).
We compare our ReMeDi with existing methods (\cref{sec: exp comparison}) and validate the proposed components through controlled ablation studies (\cref{sec: exp ablation}).
For more details and results, please refer to the supplement.

\subsection{Experimental Setup}
\label{sec: exp setup}

\mypara{Test set with GT material maps.}
Our synthetic test set, \synset{}, contains 365 high-quality objects curated from Objaverse-XL~\cite{deitke2023objaversexl}, each rendered from one view under one lighting condition, with GT base color, metalness, roughness, and normal maps at $1024 \times 1024$.
The fraction of metal pixels in the foreground varies widely across images: 80 images have less than 1\% metal pixels, and 93 have more than 50\%.
We exclude assets appearing in the available training-asset lists of the compared baselines, and hold out the selected test assets from our own training data, so that all comparing methods should not have seen these data.

\mypara{In-the-wild test set.}
We additionally curate \wildset{}, an in-house set of high-quality in-the-wild images without GT materials, for qualitative comparison.
For relighting, all methods share normals estimated using MoGe-3~\cite{kong2026moge3}.

\mypara{Metrics.}
We report metrics for both material maps and relit images, which show how the material maps work together.
For base color, we report PSNR, scale-invariant PSNR (si-PSNR, where the prediction is aligned to the GT with per-channel least squares), SSIM~\cite{wang2004image}, and LPIPS~\cite{zhang2018unreasonable}; for metalness, we report RMSE and classification accuracy at a threshold of 0.5 on foreground pixels whose GT metalness is below $\tau^\text{d}$ or above $\tau^\text{m}$; and for roughness, we report RMSE.
For relit results, we render the predicted and GT material maps with the same single-view-based renderer and GT normals, under a neutral-color environment map that is rotated in 36 steps about the viewing axis and, separately, about the world vertical axis, resulting in 72 relit frames per object per method.
Then we measure the PSNR, SSIM, and LPIPS between the rendered images and GT.

\mypara{Scope.}
We report metrics on the whole foreground (\emph{Overall}) and restricted to metal pixels (\emph{Metal}, GT metalness above $\tau^\text{m}$) and dielectric pixels (\emph{Dielectric}, below $\tau^\text{d}$).
Pixel-wise errors are evaluated within the mask of the scope.
SSIM and LPIPS are computed on the whole image, where all pixels outside the mask of the scope are replaced by the mean GT value over these pixels, in both the prediction and the GT.
Each metric is computed per image and then averaged over the images, where a restricted scope only counts the images that contain at least 5\% such pixels.

\subsection{Comparison with Existing Methods}
\label{sec: exp comparison}

\mypara{Baselines.}
We compare with methods that estimate pixel-aligned metallic--roughness maps from a single image with available model weights, including RGB$\leftrightarrow$X~\cite{zeng2024rgbx}, Intrinsic Image Diffusion (IID)~\cite{kocsis2024intrinsic}, IDArb~\cite{li2025idarb}, SuperMat~\cite{hong2025supermat}, Marigold-IID~\cite{ke2025marigold}, and the inverse renderer of DiffusionRenderer~\cite{liang2025diffusionrenderer}, which is a video generator and applied to single frames.
We run each baseline with its released checkpoint and default inference settings.

\mypara{Quantitative comparison.}
\cref{tab: comparison} reports the overall metrics on \synset{}.
As shown in \cref{tab: comparison}, ReMeDi outperforms all baselines on every metric.
Compared with the strongest baseline, SuperMat~\cite{hong2025supermat}, ReMeDi improves the PSNR of base color by 2.86\,dB and reduces its LPIPS by 57.6\%.
The RMSEs of metalness and roughness are also reduced by 25.9\% and 10.5\%, respectively.
These gains carry over to relighting, where PSNR is improved by 2.74\,dB and LPIPS drops from 0.065 to 0.031, indicating that the estimated maps combine into a more faithful appearance.

\mypara{Qualitative comparison.}
\cref{fig: comparison} shows the estimated base color and metalness by the two strongest baselines~\cite{hong2025supermat, li2025idarb} and our ReMeDi, together with the relit results, on both \synset{} and \wildset{}.
On the photosniper (left), IDArb and SuperMat estimate base color on its silver gunstock that is too dark and does not match the medium or high metalness, so that the body is rendered too dark, whereas ReMeDi correctly estimates the materials closest to the GT.
On the diving helmet (right), IDArb combines a bright base color with scattered metalness, which makes the helmet appear overly bright and noisy when relit, whereas SuperMat and ReMeDi produce plausible results that give metallic appearances.
On both examples, the base color of ReMeDi matches its metalness, leading to relit results that faithfully preserve the appearance of the input.


\subsection{Ablation Studies}
\label{sec: exp ablation}

\begin{table*}[t]
\centering
\caption{
Ablation study of the design choices in ReMeDi on \synset{}.
The ``variant'' column uses similar notations as \cref{tab: related work}.
\emph{Overall} is computed on the whole foreground, while \emph{Metal} and \emph{Dielectric} restrict every metric to metal and dielectric pixels, respectively.
Within each scope and metric, the best, second-best, and third-best results are in \textbf{bold}, \underline{underlined}, and \textit{italic}, where \notranked{gray} entries are not ranked.
}
\vspace{-6pt}
\label{tab: ablation}
\small
\setlength{\tabcolsep}{3pt}
\resizebox{\textwidth}{!}{
\begin{tabular}{@{}lc l cccc cc c ccc@{}}
\toprule
& & & \multicolumn{4}{c}{Base color} & \multicolumn{2}{c}{Metalness} & Roughness & \multicolumn{3}{c}{Relighting} \\
\cmidrule(lr){4-7} \cmidrule(lr){8-9} \cmidrule(lr){10-10} \cmidrule(lr){11-13}
 & Variant (from input image $\vI$) & Scope & PSNR$\uparrow$ & si-PSNR$\uparrow$ & SSIM$\uparrow$ & LPIPS$\downarrow$ & RMSE$\downarrow$ & Acc.$\uparrow$ & RMSE$\downarrow$ & PSNR$\uparrow$ & SSIM$\uparrow$ & LPIPS$\downarrow$ \\
\midrule
\multirow{2}{*}{(A)} & \multirow{2}{*}{\variant{$\to \vC + \vM + \vR$}{(W/o reparam., joint)}} & \emph{Overall} & \textit{19.98} & \textit{22.39} & \underline{0.938} & \textbf{0.052} & \underline{0.275} & \underline{0.830} & \textit{0.215} & \textit{20.46} & \textit{0.949} & \underline{0.044} \\
 &  & \emph{Metal} & 17.07 & 21.40 & 0.946 & 0.079 & \underline{0.523} & \underline{0.513} & 0.166 & 17.62 & 0.960 & 0.049 \\
\cmidrule(){1-13}
\multirow{2}{*}{(B)} & \multirow{2}{*}{\variant{$\to \vM \to \vC + \vR$}{(W/o reparam., metal-cond.)}} & \emph{Overall} & \underline{20.10} & \underline{22.44} & \textbf{0.939} & \textbf{0.052} & \textbf{0.237} & \textbf{0.874} & \textbf{0.204} & \underline{20.72} & \underline{0.950} & \textbf{0.043} \\
 &  & \emph{Metal} & \textit{17.81} & 21.67 & \textit{0.949} & \textit{0.076} & \textbf{0.404} & \textbf{0.673} & 0.154 & \textit{18.02} & \textit{0.961} & \textit{0.046} \\
\cmidrule(){1-13}
\multirow{3}{*}{(C)} & \multirow{3}{*}{{\variantthree{$\to \vC^{\mathrm{d}} + \vR^{\mathrm{d}} + \vC^{\mathrm{m}} + \vR^{\mathrm{m}} \,|\, \vM$}{$\Rightarrow \vC + \vR$}{(W/ reparam., single network)}}} & \emph{Overall} & 19.37 & 21.76 & \textit{0.933} & \underline{0.058} & \textbf{0.237} & \textbf{0.874} & 0.232 & 19.81 & 0.944 & \textit{0.047} \\
 &  & \emph{Metal} & 17.37 & \underline{22.11} & 0.947 & 0.078 & \textbf{0.404} & \textbf{0.673} & \textit{0.151} & 17.18 & 0.957 & 0.050 \\
 &  & \emph{Dielectric} & \textit{21.18} & \underline{25.69} & \textit{0.947} & \textit{0.064} & \textbf{0.099} & \textbf{0.955} & \textit{0.263} & \textit{22.08} & \underline{0.963} & \textit{0.043} \\
\midrule
\multirow{3}{*}{(D)} & \multirow{3}{*}{\variantthree{$\forkto \vC^{\mathrm{d}} + \vR^{\mathrm{d}} \,|\, \vC^{\mathrm{m}} + \vR^{\mathrm{m}} \,|\, \vM $}{$\Rightarrow \vC + \vR$}{(\textbf{Our full model})}} & \emph{Overall} & \textbf{20.24} & \textbf{22.52} & \textbf{0.939} & \textbf{0.052} & \textbf{0.237} & \textbf{0.874} & \underline{0.212} & \textbf{20.90} & \textbf{0.951} & \textbf{0.043} \\
 &  & \emph{Metal} & \underline{18.40} & \textit{22.08} & \underline{0.950} & \underline{0.074} & \textbf{0.404} & \textbf{0.673} & \underline{0.139} & \underline{18.76} & \underline{0.964} & \underline{0.044} \\
 &  & \emph{Dielectric} & \underline{22.08} & \textit{25.66} & \underline{0.954} & \underline{0.056} & \textbf{0.099} & \textbf{0.955} & \textbf{0.237} & \underline{22.73} & \textbf{0.967} & \underline{0.040} \\
\cmidrule(){1-13}
\multirow{2}{*}{(D-a)} & \multirow{2}{*}{\variant{$\to \vC^{\mathrm{d}} + \vR^{\mathrm{d}}$, with $\vM$ set to $\mathbf{0}$}{(Our dielectric hypothesis)}} & \emph{Overall} & 18.68 & 21.14 & 0.931 & \textit{0.059} & \notranked{0.447} & \notranked{0.676} & 0.219 & 20.19 & 0.947 & \textit{0.047} \\
 &  & \emph{Dielectric} & \textbf{22.72} & \textbf{26.08} & \textbf{0.956} & \textbf{0.053} & \notranked{0.026} & \notranked{1.000} & \underline{0.238} & \textbf{23.02} & \textbf{0.967} & \textbf{0.039} \\
\cmidrule(){1-13}
\multirow{2}{*}{(D-b)} & \multirow{2}{*}{\variant{$\to \vC^{\mathrm{m}} + \vR^{\mathrm{m}}$, with $\vM$ set to $\mathbf{1}$}{(Our metal hypothesis)}} & \emph{Overall} & 17.08 & 20.50 & 0.921 & 0.071 & \notranked{0.767} & \notranked{0.324} & 0.220 & 18.59 & 0.936 & 0.056 \\
 &  & \emph{Metal} & \textbf{20.15} & \textbf{23.64} & \textbf{0.959} & \textbf{0.065} & \notranked{0.053} & \notranked{1.000} & \textbf{0.125} & \textbf{19.52} & \textbf{0.969} & \textbf{0.039} \\
\bottomrule
\end{tabular}
}
\vspace{-11pt}
\end{table*}

The ablation study is conducted with an output resolution of $512\times 512$, an increased batch size by $4\times$, and a reduced training iteration count from 30K to 10K.
All ablated variants, including the retrained full model, share the same backbone, training data, and schedule otherwise.

\begin{figure}[!t]
\centering
\includegraphics[width=0.97\columnwidth]{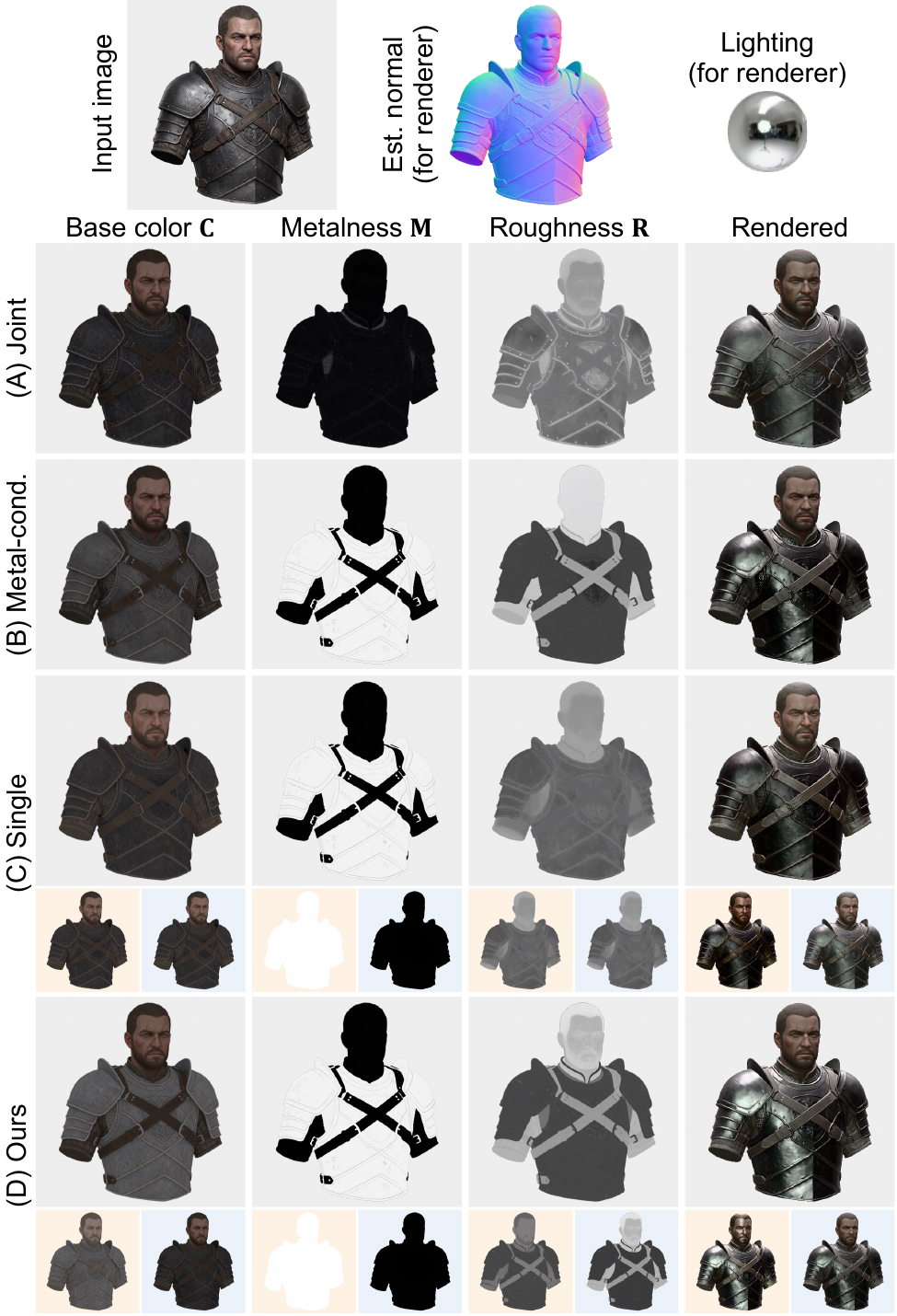}
\caption{
Qualitative ablation results on \wildset{}. 
For (C) and (D), the small images below show the maps under the metal (orange) and the dielectric (blue) hypothesis with fixed metalness.
}
\vspace{-10pt}
\label{fig: ablation}
\end{figure}

\mypara{Ablated variants.}
The variants are listed in \cref{tab: ablation}.
Variants (A) and (B) estimate base color directly with one network for both metals and dielectrics, either jointly with metalness (for (A)) or conditioned on the metalness from the same predictor as ReMeDi (for (B)).
Variant (C) keeps the dual hypotheses of material maps and the hypothesis-specific supervision, but produces them with a single network.
Variant (D) is our full model, where two hypotheses are combined per pixel by the predicted metalness, as described.
Rows (D-a) and (D-b) report the dielectric and the metal hypothesis of (D) alone, with a constant metalness of $0$ and $1$, which shows how each hypothesis contributes.

\mypara{Results.}
\cref{fig: ablation} provides visual comparisons while \cref{tab: ablation} shows quantitative results.
Although (A) and (B) usually produce base color that matches their metalness, this match is not ensured.
In \cref{fig: ablation}, (B) produces a dark base color on the armor even where its given metalness is high, so the armor renders much darker than the input, while (A) estimates inaccurate metalness and roughness so that the rendered armor loses its metallic appearance.
As shown in the small images under each estimated material map, we observe that (C) produces nearly identical base color and roughness for the two opposite hypotheses, suggesting that the single network ignores which hypothesis it is asked to follow and confirming the necessity of separate networks.
In contrast, the two hypotheses of (D) differ clearly, and the blended base color matches the metalness.
In \cref{tab: ablation}, the separate hypotheses, (D-a) and (D-b), achieve the best performance on their respective pixels (dielectric hypothesis on dielectric pixels and vice versa).
As a result, the blended (D) outperforms other variants quantitatively, especially on metal regions, where (D) achieves over +0.6\,dB improvements in base color and relighting PSNR.
\section{Conclusion}
\label{sec: conclusion}
We present ReMeDi, which estimates metallic--roughness material maps from a single image while enforcing the intended metalness of base color to match the used metalness.
ReMeDi reparameterizes the base color as a metalness-weighted blend of its two variants under the dielectric and the metal hypothesis, and estimates the two variants with separate networks, each supervised only where its hypothesis holds and thus following its hypothesis at every pixel.
Experiments show that ReMeDi not only removes the mismatch between base color and metalness but also improves material map accuracy and relighting fidelity.

\mypara{Limitations.}
Matching the base color to the estimated metalness does not correct an erroneous metalness, under which the surface is still rendered as the wrong kind of material.
Moreover, the three networks require more computation than a single one, although they can run in parallel without increasing latency.



\makeatletter\global\@colroom\@colht\global\vsize\@colht\makeatother
{
    \small
    \bibliographystyle{ieeenat_fullname}
    \bibliography{main}
}

\clearpage
\maketitlesupplementary
\appendix

This supplement provides the details and results that are omitted from the main paper.
\cref{sec: appendix data} describes how the rendered test set \synset{} is constructed, including its material distribution and its separation from the training data.
\cref{sec: appendix metrics} details the evaluation protocol, including the metrics, the renderer for relighting, and the inference settings of the baselines.
\cref{sec: appendix implementation} gives further implementation details of ReMeDi.
\cref{sec: appendix results} reports the complete quantitative results of the comparison and the ablation study on every scope.
\cref{sec: appendix qualitative} shows more qualitative comparisons with existing methods and between the ablated variants.
The attached video additionally presents an overview of ReMeDi and animated relighting results.
\textbf{We strongly recommend watching the attached video first.}

\section{Rendered Test Set Construction}
\label{sec: appendix data}

\mypara{Material distribution.}

\begin{figure}[t]
\centering
\includegraphics[width=\linewidth]{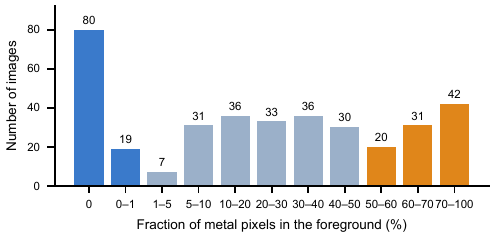}
\caption{Distribution of the fraction of metal pixels in the foreground over the 365 images of \synset{}.}
\label{fig: metal fraction}
\end{figure}

\synset{} contains 365 objects, with one selected view and one input lighting condition per object.
\cref{fig: metal fraction} shows the distribution of the fraction of metal pixels in the foreground of each image.
It covers the whole range from purely dielectric to almost purely metal objects.
80 images contain no metal pixels, 99 contain less than 1\%, 93 contain at least 50\%, and 37 contain more than 75\%, with a median of 21.5\%.
Overall, $27.1\%$ of the $93.7$M foreground pixels of the test set are metal (\ie, those with GT metalness above $0.8$, as defined in \cref{sec: exp setup}), $68.0\%$ are dielectric (with GT metalness below $0.2$), and the remaining $4.9\%$ are intermediate.

\mypara{Candidate assets and views.}
We choose 32,095 asset identifiers from our in-house rendered data, and exclude identifiers in the released IDArb \cite{li2025idarb} training list and Objaverse-LVIS used by DiffusionRenderer \cite{liang2025diffusionrenderer}, leaving 28,727 candidates.
Identifiers are matched by their 32-character hexadecimal base identifier so that variants with suffixes are also excluded.
Restricting the candidates to publicly sourced Objaverse assets yields 2,348 objects, of which 1,241 match Objaverse 1.0 \cite{deitke2023objaverse} identifiers and 1,107 come from the GitHub, Thingiverse, and Smithsonian extensions of Objaverse-XL \cite{deitke2023objaversexl}.
The source batches render 38 cameras per object at $2048\times2048$, including 14 fixed and 24 random cameras.
Using each object identifier as a random seed, we select five candidate views, including fixed and random views, for inspection.

\mypara{Candidate selection.}
We define metal pixels by $\vM>0.8$, dielectric pixels by $\vM<0.2$, and foreground by alpha of at least 128 in the original 8-bit mask.
To cover a wide range of metal fractions, candidates were drawn with pooled foreground counts across the candidate views, including intermediate-metalness pixels in the denominator; these counts are used only for selection, and all metric masks are computed from the selected view.

\mypara{Quality control.}
We automatically reject assets with near-constant metalness and roughness defaults, more than 20\% intermediate-metalness pixels, missing base color textures, implausibly dark metallic base color, or detected duplicates.
View-level checks reject tiny or cropped foregrounds.
One author inspects the candidate inputs and all GT components, keeping at most one view--lighting pair per object.
The usable input light types are area light, environment light, white point light, and weakly colored point light; heavily overexposed inputs are excluded from selection.

\mypara{Training separation.}
For deduplication, we compare orientation-independent descriptors of dimensions, radial geometry, base color, and metalness/roughness across fixed views; candidate duplicate groups are consolidated before selection.
The holdout list contains the 365 selected identifiers and 49 detected copies, totaling 414 identifiers.
All these assets are removed from the training data of every ReMeDi module and every ablation variant, which are all trained after the removal.

\section{Evaluation Protocol}
\label{sec: appendix metrics}

\mypara{Pixelwise metrics and aggregation.}
PSNR and RMSE use the GT foreground mask.
Each image contributes its own normalized error rather than contributing in proportion to its number of pixels.
\emph{Overall} averages the per-image scores over all 365 images; \emph{Metal} and \emph{Dielectric} average them over the images whose scope covers at least $5\%$ of their foreground and at least $100$ pixels.
For relighting, scores are first averaged across the evaluation lighting conditions within an image, then across images.
si-PSNR is evaluated on the foreground pixels only: each channel of the prediction is rescaled by its least-squares factor $\sum pg/\sum p^2$ before the error is measured, the three per-channel errors are averaged, and the result is converted to decibels on the same $[0,1]$ intensity range as PSNR. No clipping is applied to either the prediction or the GT, and images whose foreground covers fewer than $100$ pixels are skipped.
Metalness classification is evaluated on foreground pixels with GT $\vM<0.2$ or $\vM>0.8$; intermediate values are excluded from this classification score.

\mypara{Spatial metrics.}
LPIPS and SSIM use a common fixed background for predicted and GT images, with identical framing and resolution, and are computed per image and averaged over the images of the scope.
For \emph{Metal} and \emph{Dielectric}, every pixel outside the scope is replaced, in both images, by the mean GT value over these pixels.
SSIM uses an $11\times11$ Gaussian window with $\sigma=1.5$~\cite{wang2004image}; LPIPS uses the AlexNet~\cite{krizhevsky2012imagenet} backbone.

\mypara{Relighting.}
Predicted and GT materials are passed to the same single-view-based renderer with identical normals, view directions, lighting, and any auxiliary geometry or occlusion inputs.
The GT-material render instead of the original input image is used as the evaluation reference to keep the only difference to be the used materials.
For in-the-wild examples, all methods instead share the same estimated normals from the input, and no GT reference is available.
The renderer shades every pixel independently under an orthographic view, without cast shadows or interreflections.
The BRDF is a Lambertian diffuse lobe with albedo $(1-\vM)\,\vC$ plus a GGX specular lobe~\cite{walter2007microfacet} with $\alpha = \vR^2$, separable Smith visibility, and Schlick Fresnel with $F_0 = 0.04\,(1-\vM) + \vC\,\vM$~\cite{karis2013real}; base color is decoded from sRGB to linear, and $\vR$ is clamped to at least $0.02$.
Lighting is the studio environment map \emph{studio\_kominka\_01}~\cite{polyhaven_kominka}, white-balanced so that its mean radiance is neutral and yawed so that its dominant light lies in the upper front.
Specular lighting uses the split-sum approximation~\cite{karis2013real}: the environment is prefiltered with GGX importance sampling~\cite{krivanek2008realtime} at ten roughness levels and combined with a DFG lookup table computed for the same BRDF, looked up along the dominant specular direction~\cite{lagarde2014moving}. Diffuse lighting uses the exact cosine convolution of the environment.
Energy lost to single scattering is restored with multiple-scattering compensation~\cite{fdezaguera2019multiple}.
The shared GT ambient occlusion attenuates the diffuse term with a multi-bounce approximation~\cite{jimenez2016practical} and the specular term with specular occlusion~\cite{lagarde2014moving}.
The environment is rotated in 36 steps of $10^\circ$ about the viewing axis and, in a second sequence, about the world vertical axis, which is obtained from the rendering camera of each test view.
Exposure is metered on a white Lambertian card facing the camera, averaged over the frames of a sequence, so that it is independent of the object and of the method; the linear radiance is scaled by this exposure, tone-mapped with the Khronos PBR Neutral operator~\cite{khronos_pbr_neutral}, and quantized to 8-bit sRGB, with no further adjustment.
GT material maps are exported with premultiplied alpha and are un-premultiplied before shading.
Relighting scores are averaged over the $72$ frames of the two sequences.

\mypara{Baseline inference.}
All baselines use their released checkpoints and default inference settings, with the input composited over the background each method expects and resized to its native resolution.
SuperMat~\cite{hong2025supermat} runs at $512\times512$ with a single denoising step on a gray background of value $0.5$.
RGB$\leftrightarrow$X~\cite{zeng2024rgbx} runs at $512\times512$ on a white background with 50 steps for each of its five per-channel prompts; its released metalness weights differ from those described in the paper.
IID~\cite{kocsis2024intrinsic} reads roughness and metalness from the first and second channels of its material output, and its albedo output is already in sRGB and is used without gamma correction.
IDArb~\cite{li2025idarb} runs at $512\times512$ on a white background with 50 steps.
Marigold-IID~\cite{ke2025marigold} runs at $768\times768$ with 4 steps and an ensemble of 10 samples.
DiffusionRenderer~\cite{liang2025diffusionrenderer} runs its inverse renderer on a single frame at $512\times512$ with 20 steps on a black background.
All outputs are resized back to $1024\times1024$ and masked by the input alpha before evaluation, and all methods are relit with the same normals and lighting.
\begin{table*}[t]
\centering
\caption{Complete quantitative comparison results on \synset{} with GT materials.
\emph{Overall} is computed on the whole foreground, while \emph{Metal} and \emph{Dielectric} restrict every metric to metal and dielectric pixels, respectively.
The last two rows apply the dielectric and the metal network of ReMeDi alone, with a constant metalness of $0$ and $1$.
Within each scope and metric, the best, second-best, and third-best results are in \textbf{bold}, \underline{underlined}, and \textit{italic}, where \notranked{gray} entries do not participate in ranking.}
\label{tab: comparison complete}
\small
\setlength{\tabcolsep}{3pt}
\begin{tabular}{@{}ll cccc cc c ccc@{}}
\toprule
& & \multicolumn{4}{c}{Base color} & \multicolumn{2}{c}{Metalness} & Roughness & \multicolumn{3}{c}{Relighting} \\
\cmidrule(lr){3-6} \cmidrule(lr){7-8} \cmidrule(lr){9-9} \cmidrule(lr){10-12}
Method & Scope & PSNR$\uparrow$ & si-PSNR$\uparrow$ & SSIM$\uparrow$ & LPIPS$\downarrow$ & RMSE$\downarrow$ & Acc.$\uparrow$ & RMSE$\downarrow$ & PSNR$\uparrow$ & SSIM$\uparrow$ & LPIPS$\downarrow$ \\
\midrule
\multirow{3}{*}{RGB$\leftrightarrow$X~\cite{zeng2024rgbx}} & \emph{Overall} & 13.16 & 16.85 & 0.881 & 0.120 & 0.599 & 0.388 & 0.258 & 14.56 & 0.897 & 0.091 \\
 & \emph{Metal} & 12.98 & 16.84 & 0.941 & 0.088 & 0.425 & 0.677 & 0.210 & 13.96 & 0.948 & 0.059 \\
 & \emph{Dielectric} & 14.17 & 20.52 & 0.903 & 0.110 & 0.614 & 0.288 & 0.280 & 15.55 & 0.922 & 0.080 \\\cmidrule(){1-12}
\multirow{3}{*}{IID~\cite{kocsis2024intrinsic}} & \emph{Overall} & 14.09 & 17.89 & 0.893 & 0.121 & \textit{0.399} & 0.690 & 0.262 & 15.52 & 0.913 & 0.087 \\
 & \emph{Metal} & 16.09 & 20.18 & 0.949 & 0.080 & 0.648 & 0.142 & 0.153 & 15.61 & 0.955 & 0.053 \\
 & \emph{Dielectric} & 13.47 & 21.19 & 0.910 & 0.117 & \textit{0.265} & \underline{0.931} & 0.296 & 15.66 & 0.933 & 0.080 \\\cmidrule(){1-12}
\multirow{3}{*}{IDArb~\cite{li2025idarb}} & \emph{Overall} & 16.60 & 18.53 & 0.905 & 0.092 & 0.473 & 0.628 & 0.273 & 17.57 & 0.914 & 0.074 \\
 & \emph{Metal} & 15.12 & 17.96 & 0.947 & 0.076 & 0.604 & 0.396 & 0.207 & 15.85 & 0.954 & 0.050 \\
 & \emph{Dielectric} & 17.82 & 22.36 & 0.927 & 0.084 & 0.409 & 0.702 & 0.302 & 18.81 & 0.939 & 0.063 \\\cmidrule(){1-12}
\multirow{3}{*}{SuperMat~\cite{hong2025supermat}} & \emph{Overall} & \textit{18.47} & \textit{20.62} & 0.920 & 0.094 & \underline{0.266} & \underline{0.847} & \textit{0.191} & \textit{19.60} & 0.937 & 0.065 \\
 & \emph{Metal} & \textit{17.28} & 20.71 & 0.955 & 0.076 & \underline{0.283} & \underline{0.822} & \textit{0.148} & \textit{18.08} & 0.965 & 0.045 \\
 & \emph{Dielectric} & \textit{19.64} & \textit{23.50} & \textit{0.938} & 0.084 & \underline{0.255} & \textit{0.825} & \textit{0.204} & \textit{21.22} & \textit{0.957} & \textit{0.055} \\\cmidrule(){1-12}
\multirow{3}{*}{Marigold-IID~\cite{ke2025marigold}} & \emph{Overall} & 15.55 & 18.95 & 0.905 & 0.121 & 0.504 & 0.450 & 0.222 & 17.00 & 0.925 & 0.080 \\
 & \emph{Metal} & 17.16 & 20.26 & 0.956 & 0.079 & \textbf{0.281} & \textbf{0.936} & 0.219 & 16.05 & 0.959 & 0.052 \\
 & \emph{Dielectric} & 14.98 & 22.47 & 0.917 & 0.113 & 0.615 & 0.177 & 0.220 & 17.65 & 0.944 & 0.071 \\\cmidrule(){1-12}
\multirow{3}{*}{DiffusionRenderer~\cite{liang2025diffusionrenderer}} & \emph{Overall} & 15.31 & 18.55 & 0.901 & 0.105 & 0.413 & \textit{0.693} & 0.243 & 16.59 & 0.917 & 0.078 \\
 & \emph{Metal} & 14.86 & 17.66 & 0.948 & 0.084 & \textit{0.320} & 0.779 & 0.217 & 15.17 & 0.954 & 0.055 \\
 & \emph{Dielectric} & 16.17 & 21.92 & 0.922 & 0.096 & 0.434 & 0.636 & 0.256 & 18.03 & 0.942 & 0.066 \\\cmidrule(){1-12}
\multirow{3}{*}{\textbf{ReMeDi (Ours)}} & \emph{Overall} & \textbf{21.33} & \textbf{23.79} & \textbf{0.956} & \textbf{0.039} & \textbf{0.197} & \textbf{0.913} & \textbf{0.170} & \textbf{22.34} & \textbf{0.965} & \textbf{0.031} \\
 & \emph{Metal} & \underline{19.65} & \underline{22.80} & \underline{0.968} & \underline{0.051} & 0.322 & \textit{0.791} & \underline{0.141} & \textbf{19.89} & \textbf{0.974} & \textbf{0.031} \\
 & \emph{Dielectric} & \underline{23.11} & \underline{26.97} & \underline{0.969} & \underline{0.037} & \textbf{0.088} & \textbf{0.972} & \underline{0.178} & \underline{24.69} & \underline{0.980} & \underline{0.023} \\\cmidrule(){1-12}
\multirow{3}{*}{ReMeDi dielectric hypothesis} & \emph{Overall} & \underline{19.97} & \underline{22.40} & \underline{0.951} & \underline{0.042} & \notranked{0.450} & \notranked{0.675} & \underline{0.173} & \underline{21.47} & \underline{0.962} & \underline{0.034} \\
 & \emph{Metal} & 15.94 & \textit{20.89} & \textit{0.957} & \textit{0.062} & \notranked{0.967} & \notranked{0.000} & 0.156 & 17.38 & \textit{0.967} & \textit{0.039} \\
 & \emph{Dielectric} & \textbf{23.96} & \textbf{27.81} & \textbf{0.971} & \textbf{0.033} & \notranked{0.027} & \notranked{1.000} & \textbf{0.177} & \textbf{25.11} & \textbf{0.981} & \textbf{0.022} \\\cmidrule(){1-12}
\multirow{3}{*}{ReMeDi metal hypothesis} & \emph{Overall} & 15.72 & 20.10 & \textit{0.926} & \textit{0.071} & \notranked{0.767} & \notranked{0.325} & 0.223 & 17.77 & \textit{0.939} & \textit{0.053} \\
 & \emph{Metal} & \textbf{20.14} & \textbf{24.17} & \textbf{0.969} & \textbf{0.050} & \notranked{0.051} & \notranked{1.000} & \textbf{0.139} & \underline{19.50} & \underline{0.973} & \underline{0.032} \\
 & \emph{Dielectric} & 14.44 & 23.43 & 0.929 & \textit{0.082} & \notranked{0.985} & \notranked{0.000} & 0.254 & 17.41 & 0.951 & \textit{0.055} \\
\bottomrule
\end{tabular}
\end{table*}

\section{Implementation Details}
\label{sec: appendix implementation}

\mypara{Background tokens.}
We neither supervise background tokens nor pass them through the transformer by keeping only the tokens $\mathcal{K}$ that cover the object and a ring of tokens around it, and setting $w_k^h = 0$ elsewhere.
The ring is kept because the VAE encoder spreads information from the silhouette into the neighboring tokens.
Before decoding, the dropped target tokens are filled with the corresponding condition tokens.
This accelerates both training and inference, as the token sequence is shortened from $N$ to $|\mathcal{K}|$ tokens for each material map.
Moreover, as metal regions usually cover only a small part of an image, the supervision on metal tokens would be overwhelmed by the background tokens, whereas restricting the loss to $\mathcal{K}$ prevents this and leads to faster convergence.

\mypara{Text prompts.}
Each module uses a fixed text prompt that describes its task.
The dielectric and the metal module are told to assume that the entire object is made of a dielectric and a metal material, respectively, \eg, for the metal module:
\begin{quote}
\small\emph{Assume the entire object is made of metal. From the input image alone, jointly generate a physically plausible metallic-material basecolor texture map and a roughness texture map for the whole object, as if every surface were metallic. Remove all directional lighting, cast shadows, and specular highlights; keep all maps spatially aligned and consistent across parts, preserving geometry, seams, and material boundaries.}
\end{quote}
The prompt of the dielectric module replaces the first two sentences by \emph{``Assume the entire object is made of non-metallic (dielectric) material.''} and \emph{``\ldots a physically plausible non-metallic basecolor texture map \ldots, as if no surface were metallic.''}, and the metalness module asks for a metalness map where bright values represent metal surfaces.

\section{Complete Quantitative Results}
\label{sec: appendix results}

\cref{tab: comparison complete} reports the complete metric set of \cref{tab: comparison} on the \emph{Overall}, \emph{Metal}, and \emph{Dielectric} scopes, and \cref{tab: ablation complete} gives the same breakdown for the ablation variants of \cref{tab: ablation}.

\mypara{Comparison with existing methods.}
In \cref{tab: comparison complete}, ReMeDi outperforms all baselines on every metric in the \emph{Overall} scope, and on base color, roughness, and relighting in both the \emph{Metal} and the \emph{Dielectric} scope.
Compared with the strongest baseline, SuperMat, ReMeDi improves the base color PSNR by 2.37\,dB on metal pixels and by 3.47\,dB on dielectric pixels, and the relighting PSNR by 1.81\,dB and 3.47\,dB, respectively.
On metal pixels, Marigold-IID and SuperMat classify metalness more accurately than ReMeDi.
However, Marigold-IID classifies only 17.7\% of dielectric pixels correctly, \ie, it tends to predict high metalness everywhere, while ReMeDi keeps an accuracy of 97.2\% on dielectric pixels and achieves the best accuracy in the \emph{Overall} scope.
The two hypotheses of ReMeDi alone each estimate the best base color on the pixels where they hold, \ie, the dielectric hypothesis on dielectric pixels and the metal hypothesis on metal pixels, but degrade on the other kind of pixels.
The blended result stays within 0.5\,dB and 0.9\,dB of the base color PSNR of the corresponding hypothesis on metal and dielectric pixels, respectively, while being the best in the \emph{Overall} scope.
As the blended maps equal those of the metal hypothesis wherever $\vM = 1$, the remaining gap on metal pixels comes from the pixels whose estimated metalness is below $1$.

\mypara{Ablation study.}
\cref{tab: ablation complete} shows that the advantage of the full model (D) over the other variants mainly comes from metal pixels.
On metal pixels, (D) improves the base color PSNR over (A), (B), and (C) by 1.33\,dB, 0.59\,dB, and 1.03\,dB, and the relighting PSNR by 1.14\,dB, 0.74\,dB, and 1.58\,dB, respectively, and also achieves the lowest roughness RMSE among them.
As (B), (C), and (D) share the same metalness, these gains come from the base color and roughness alone.
Moreover, the gap between PSNR and si-PSNR of base color on metal pixels, which measures the error that a per-channel scale can remove, is the smallest for (D), with 3.68\,dB compared with 4.33\,dB, 3.86\,dB, and 4.74\,dB for (A), (B), and (C).
This indicates that a larger part of the error of the other variants comes from the brightness of base color, which is consistent with a base color that mismatches the metalness.
In particular, (C) achieves a si-PSNR on metal pixels comparable with (D), but a 1.03\,dB lower PSNR, which is consistent with its nearly identical hypotheses in \cref{fig: ablation}.
On dielectric pixels, (A) and (B) are slightly better than (D), by at most 0.45\,dB in base color PSNR and 0.21\,dB in relighting PSNR, while (C) is the worst among (A)--(D).
Consistent with \cref{tab: comparison complete}, the two hypotheses of (D), (D-a) and (D-b), each estimate the best base color on the pixels where they hold.

\begin{table*}[t]
\centering
\caption{Complete metric set of the ablation variants of \cref{tab: ablation}, for the \emph{Overall}, \emph{Metal}, and \emph{Dielectric} scopes of \cref{tab: ablation}.
The pipelines use the notation of \cref{tab: ablation}; the scopes and the treatment of the metrics that are not defined on a pixel scope follow \cref{tab: ablation}, and every variant is reported on all three scopes even where \cref{tab: ablation} omits one.
Within each scope and metric, the best, second-best, and third-best results are in \textbf{bold}, \underline{underlined}, and \textit{italic}, where \notranked{gray} entries do not participate in ranking.
}
\label{tab: ablation complete}
\small
\setlength{\tabcolsep}{3pt}
\resizebox{\textwidth}{!}{
\begin{tabular}{@{}lc l cccc cc c ccc@{}}
\toprule
& & & \multicolumn{4}{c}{Base color} & \multicolumn{2}{c}{Metalness} & Roughness & \multicolumn{3}{c}{Relighting} \\
\cmidrule(lr){4-7} \cmidrule(lr){8-9} \cmidrule(lr){10-10} \cmidrule(lr){11-13}
 & Variant (from input image $\vI$) & Scope & PSNR$\uparrow$ & si-PSNR$\uparrow$ & SSIM$\uparrow$ & LPIPS$\downarrow$ & RMSE$\downarrow$ & Acc.$\uparrow$ & RMSE$\downarrow$ & PSNR$\uparrow$ & SSIM$\uparrow$ & LPIPS$\downarrow$ \\
\midrule
\multirow{3}{*}{(A)} & \multirow{3}{*}{\variant{$\to \vC + \vM + \vR$}{(W/o reparam., joint)}} & \emph{Overall} & \textit{19.98} & \textit{22.39} & \underline{0.938} & \textbf{0.052} & \underline{0.275} & \underline{0.830} & \textit{0.215} & \textit{20.46} & \textit{0.949} & \underline{0.044} \\
 &  & \emph{Metal} & 17.07 & 21.40 & 0.946 & 0.079 & \underline{0.523} & \underline{0.513} & 0.166 & 17.62 & 0.960 & 0.049 \\
 &  & \emph{Dielectric} & \underline{22.53} & \underline{25.85} & \underline{0.955} & \textbf{0.053} & \textbf{0.094} & \textbf{0.973} & \underline{0.230} & \textit{22.77} & \textbf{0.968} & \textbf{0.038} \\
\cmidrule(){1-13}
\multirow{3}{*}{(B)} & \multirow{3}{*}{\variant{$\to \vM \to \vC + \vR$}{(W/o reparam., metal-cond.)}} & \emph{Overall} & \underline{20.10} & \underline{22.44} & \textbf{0.939} & \textbf{0.052} & \textbf{0.237} & \textbf{0.874} & \textbf{0.204} & \underline{20.72} & \underline{0.950} & \textbf{0.043} \\
 &  & \emph{Metal} & \textit{17.81} & 21.67 & \textit{0.949} & \textit{0.076} & \textbf{0.404} & \textbf{0.673} & 0.154 & \textit{18.02} & \textit{0.961} & \textit{0.046} \\
 &  & \emph{Dielectric} & \textit{22.21} & \textit{25.70} & \textit{0.954} & \underline{0.054} & \underline{0.099} & \underline{0.955} & \textbf{0.219} & \underline{22.94} & \textbf{0.968} & \underline{0.039} \\
\cmidrule(){1-13}
\multirow{3}{*}{(C)} & \multirow{3}{*}{{\variantthree{$\to \vC^{\mathrm{d}} + \vR^{\mathrm{d}} + \vC^{\mathrm{m}} + \vR^{\mathrm{m}} \,|\, \vM$}{$\Rightarrow \vC + \vR$}{(W/ reparam., single network)}}} & \emph{Overall} & 19.37 & 21.76 & \textit{0.933} & \underline{0.058} & \textbf{0.237} & \textbf{0.874} & 0.232 & 19.81 & 0.944 & \textit{0.047} \\
 &  & \emph{Metal} & 17.37 & \underline{22.11} & 0.947 & 0.078 & \textbf{0.404} & \textbf{0.673} & \textit{0.151} & 17.18 & 0.957 & 0.050 \\
 &  & \emph{Dielectric} & 21.18 & 25.69 & 0.947 & 0.064 & \underline{0.099} & \underline{0.955} & 0.263 & 22.08 & \textit{0.963} & 0.043 \\
\cmidrule(){1-13}
\multirow{3}{*}{(D)} & \multirow{3}{*}{\variantthree{$\forkto \vC^{\mathrm{d}} + \vR^{\mathrm{d}} \,|\, \vC^{\mathrm{m}} + \vR^{\mathrm{m}} \,|\, \vM $}{$\Rightarrow \vC + \vR$}{(\textbf{Our full model})}} & \emph{Overall} & \textbf{20.24} & \textbf{22.52} & \textbf{0.939} & \textbf{0.052} & \textbf{0.237} & \textbf{0.874} & \underline{0.212} & \textbf{20.90} & \textbf{0.951} & \textbf{0.043} \\
 &  & \emph{Metal} & \underline{18.40} & \textit{22.08} & \underline{0.950} & \underline{0.074} & \textbf{0.404} & \textbf{0.673} & \underline{0.139} & \underline{18.76} & \underline{0.964} & \underline{0.044} \\
 &  & \emph{Dielectric} & 22.08 & 25.66 & \textit{0.954} & \textit{0.056} & \underline{0.099} & \underline{0.955} & \textit{0.237} & 22.73 & \underline{0.967} & \textit{0.040} \\
\cmidrule(){1-13}
\multirow{3}{*}{(D-a)} & \multirow{3}{*}{\variant{$\to \vC^{\mathrm{d}} + \vR^{\mathrm{d}}$, with $\vM$ set to $\mathbf{0}$}{(Our dielectric hypothesis)}} & \emph{Overall} & 18.68 & 21.14 & 0.931 & \textit{0.059} & \notranked{0.447} & \notranked{0.676} & 0.219 & 20.19 & 0.947 & \textit{0.047} \\
 &  & \emph{Metal} & 14.26 & 20.47 & 0.934 & 0.092 & \notranked{0.965} & \notranked{0.000} & 0.160 & 16.85 & 0.957 & 0.055 \\
 &  & \emph{Dielectric} & \textbf{22.72} & \textbf{26.08} & \textbf{0.956} & \textbf{0.053} & \notranked{0.026} & \notranked{1.000} & 0.238 & \textbf{23.02} & \underline{0.967} & \underline{0.039} \\
\cmidrule(){1-13}
\multirow{3}{*}{(D-b)} & \multirow{3}{*}{\variant{$\to \vC^{\mathrm{m}} + \vR^{\mathrm{m}}$, with $\vM$ set to $\mathbf{1}$}{(Our metal hypothesis)}} & \emph{Overall} & 17.08 & 20.50 & 0.921 & 0.071 & \notranked{0.767} & \notranked{0.324} & 0.220 & 18.59 & 0.936 & 0.056 \\
 &  & \emph{Metal} & \textbf{20.15} & \textbf{23.64} & \textbf{0.959} & \textbf{0.065} & \notranked{0.053} & \notranked{1.000} & \textbf{0.125} & \textbf{19.52} & \textbf{0.969} & \textbf{0.039} \\
 &  & \emph{Dielectric} & 16.12 & 23.97 & 0.921 & 0.096 & \notranked{0.985} & \notranked{0.000} & 0.254 & 18.46 & 0.945 & 0.063 \\
\bottomrule
\end{tabular}
}
\end{table*}

\section{More Qualitative Results}
\label{sec: appendix qualitative}

For an animated comparison, please see the attached video.

\mypara{Comparison with existing methods.}

\begin{figure*}[t!]
\centering
\includegraphics[width=\textwidth]{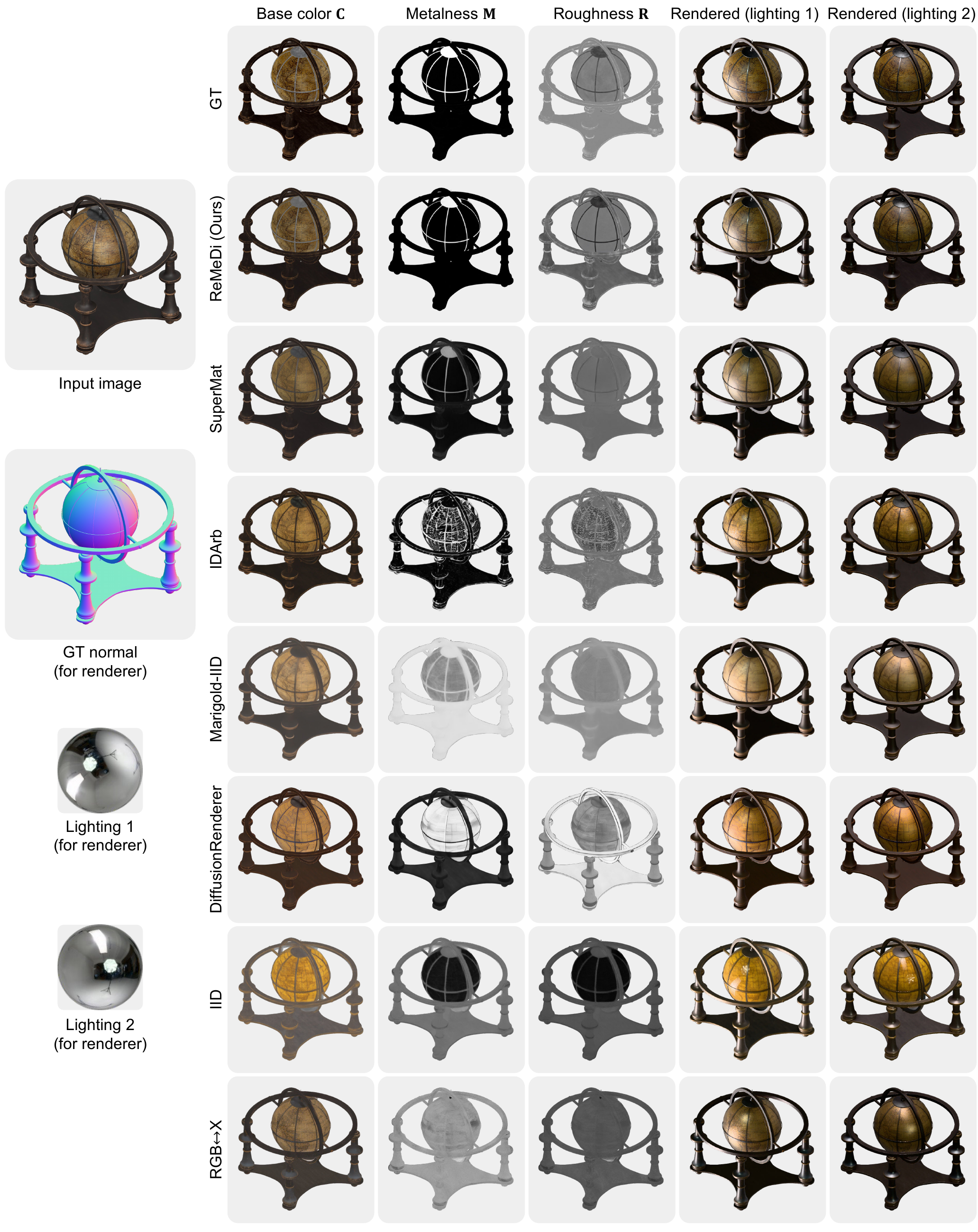}
\caption{
Qualitative comparison on \synset{} with GT.
We show the base color, metalness, roughness, and relit results using GT normals.
}
\label{fig: appendix comparison syntest 1}
\end{figure*}

\begin{figure*}[t!]
\centering
\includegraphics[width=\textwidth]{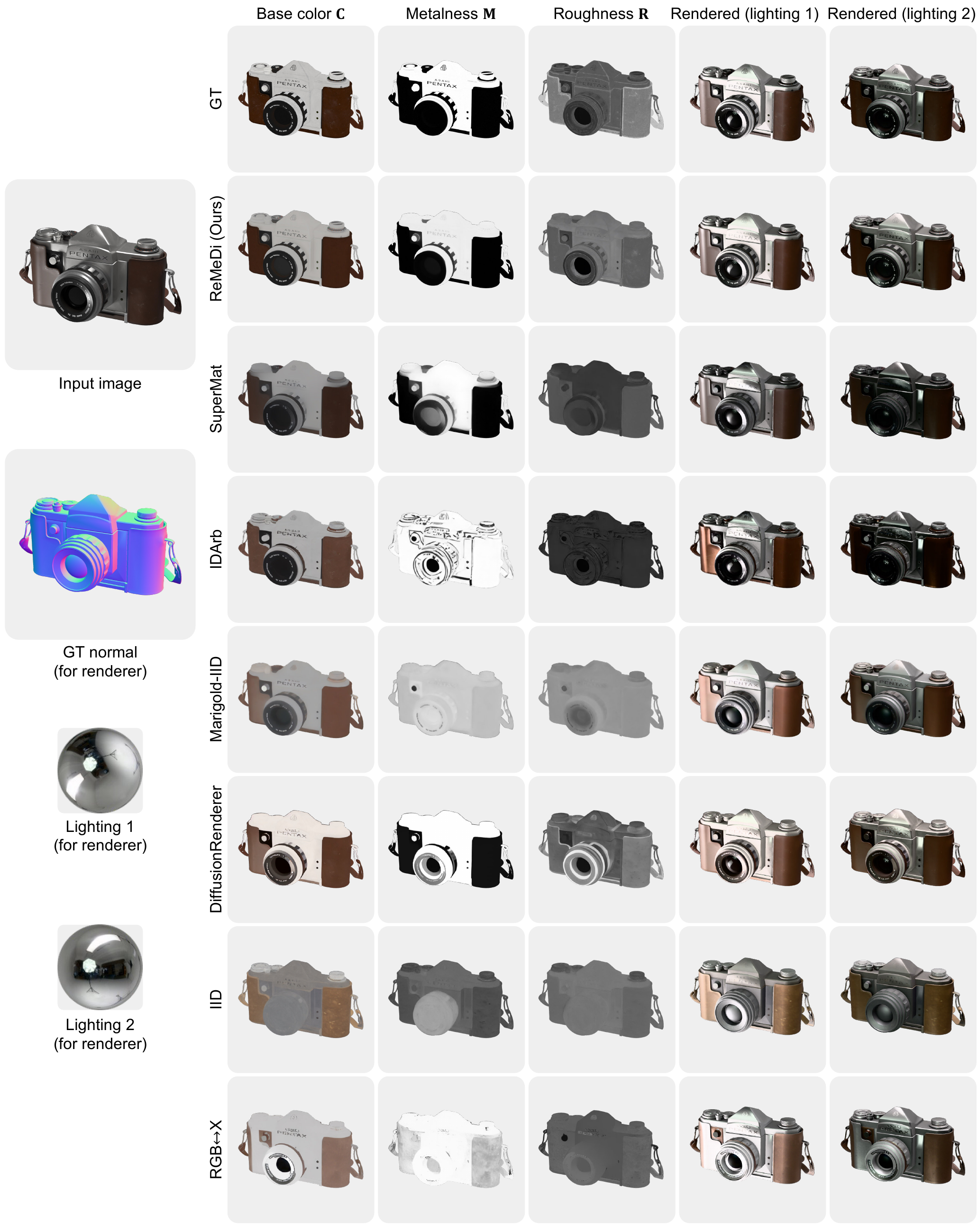}
\caption{
Qualitative comparison on \synset{} with GT.
We show the base color, metalness, roughness, and relit results using GT normals.
}
\label{fig: appendix comparison syntest 2}
\end{figure*}

\begin{figure*}[t!]
\centering
\includegraphics[width=\textwidth]{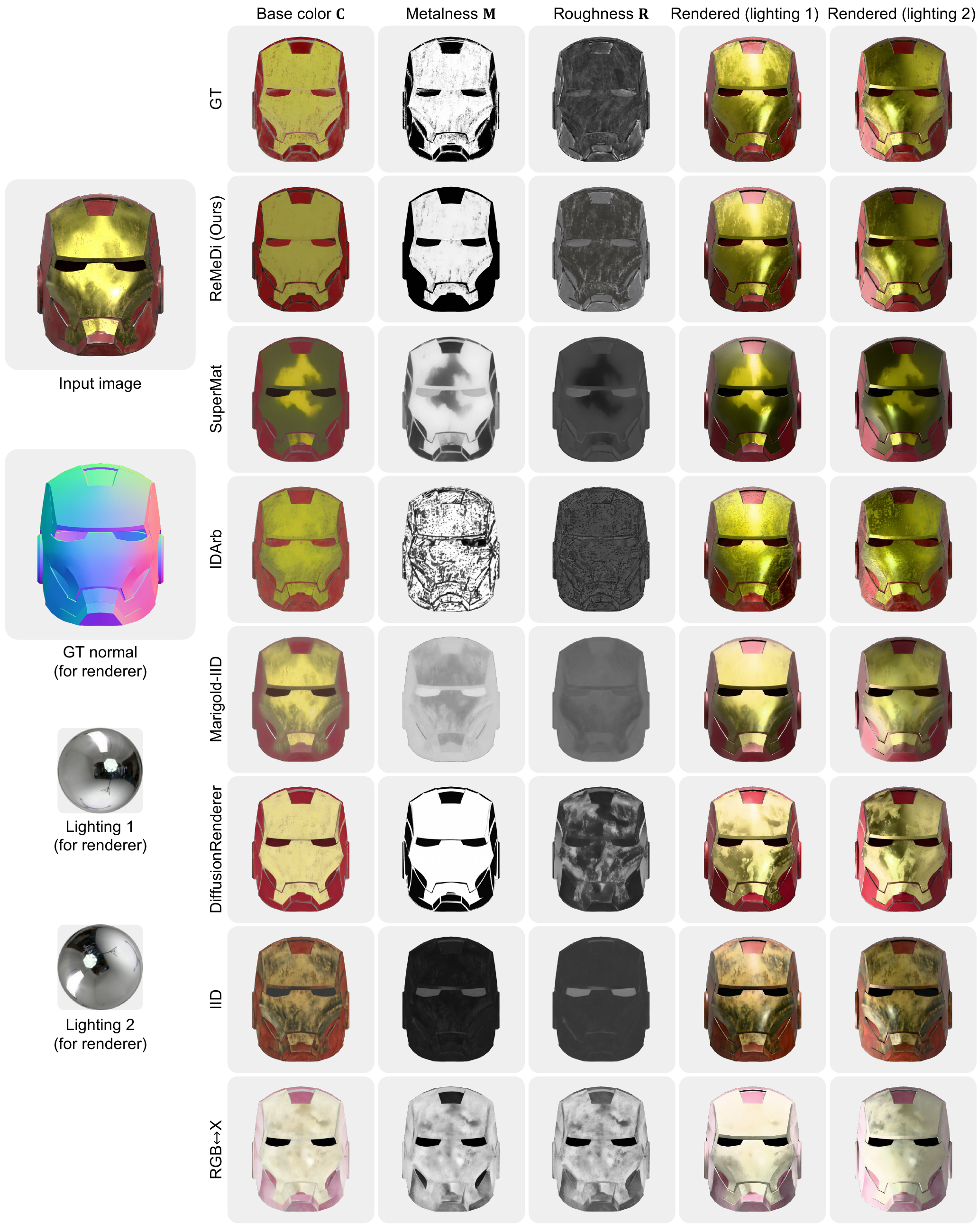}
\caption{
Qualitative comparison on \synset{} with GT.
We show the base color, metalness, roughness, and relit results using GT normals.
}
\label{fig: appendix comparison syntest 3}
\end{figure*}

\begin{figure*}[t!]
\centering
\includegraphics[width=0.99\textwidth]{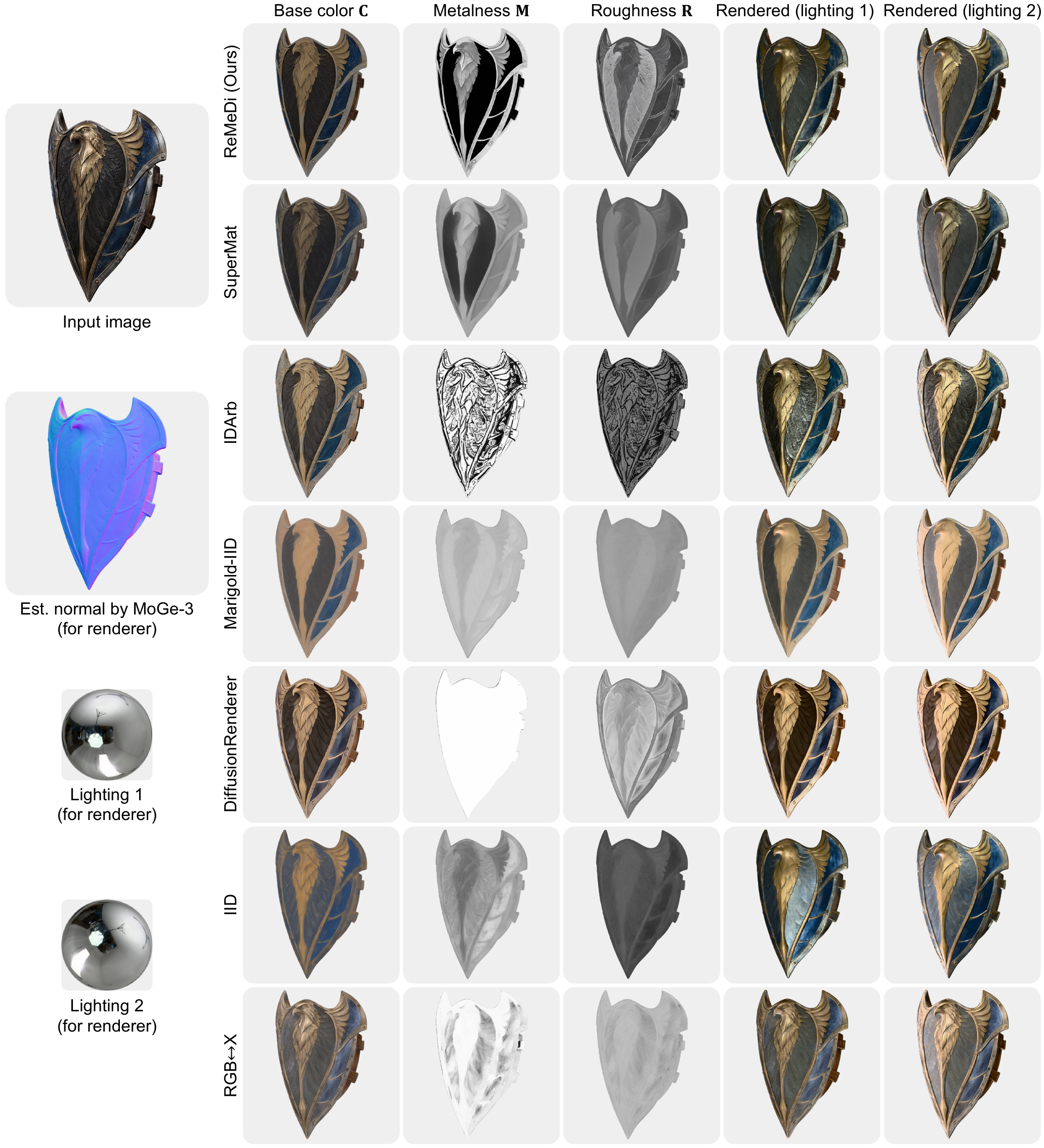}
\caption{
Qualitative comparison on \wildset{}.
We show the base color, metalness, roughness, and relit results using estimated normals.
}
\label{fig: appendix comparison wildtest 1}
\end{figure*}

\begin{figure*}[t!]
\centering
\includegraphics[width=0.99\textwidth]{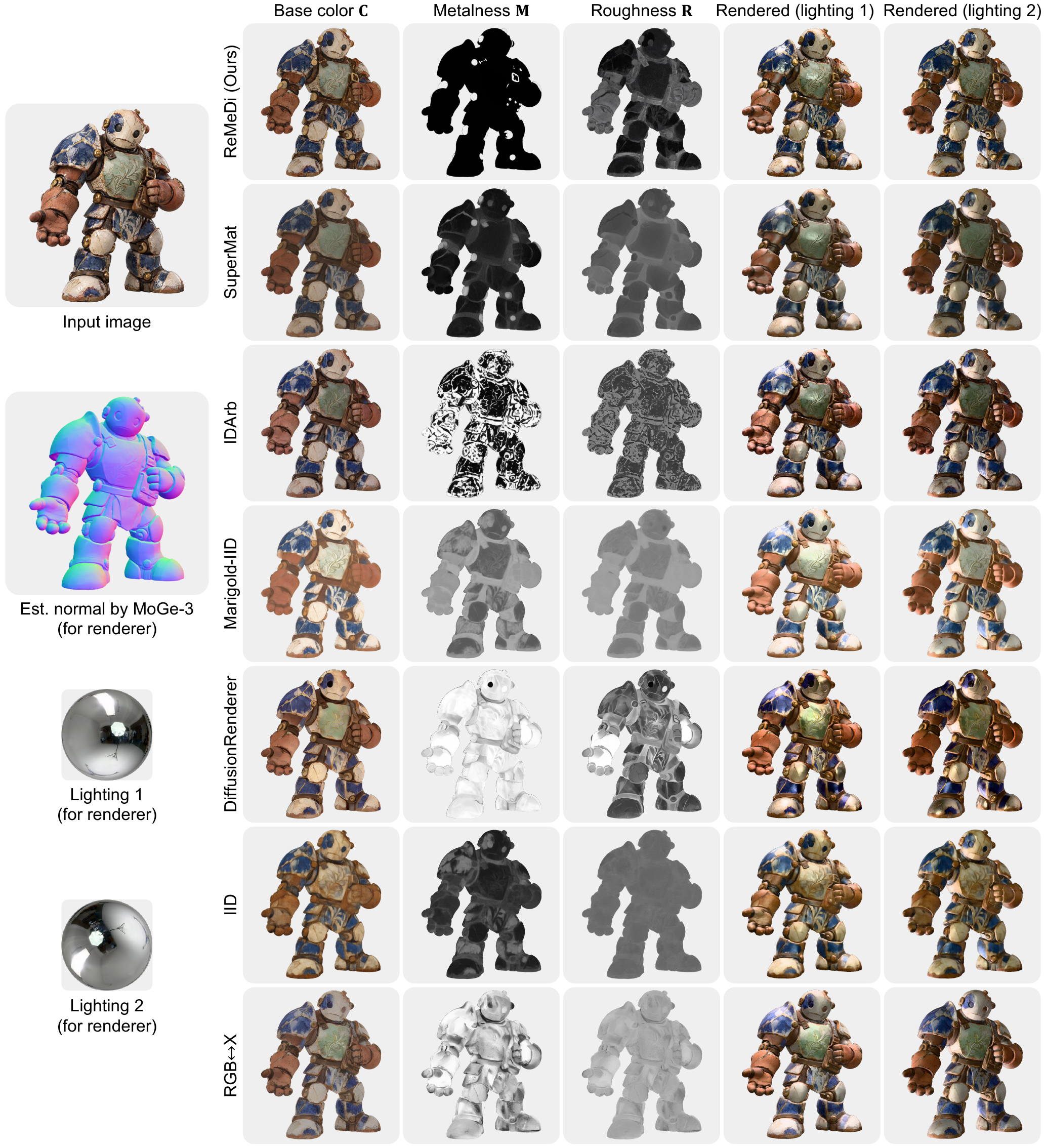}
\caption{
Qualitative comparison on \wildset{}.
We show the base color, metalness, roughness, and relit results using estimated normals.
}
\label{fig: appendix comparison wildtest 2}
\end{figure*}

\begin{figure*}[t!]
\centering
\includegraphics[width=0.99\textwidth]{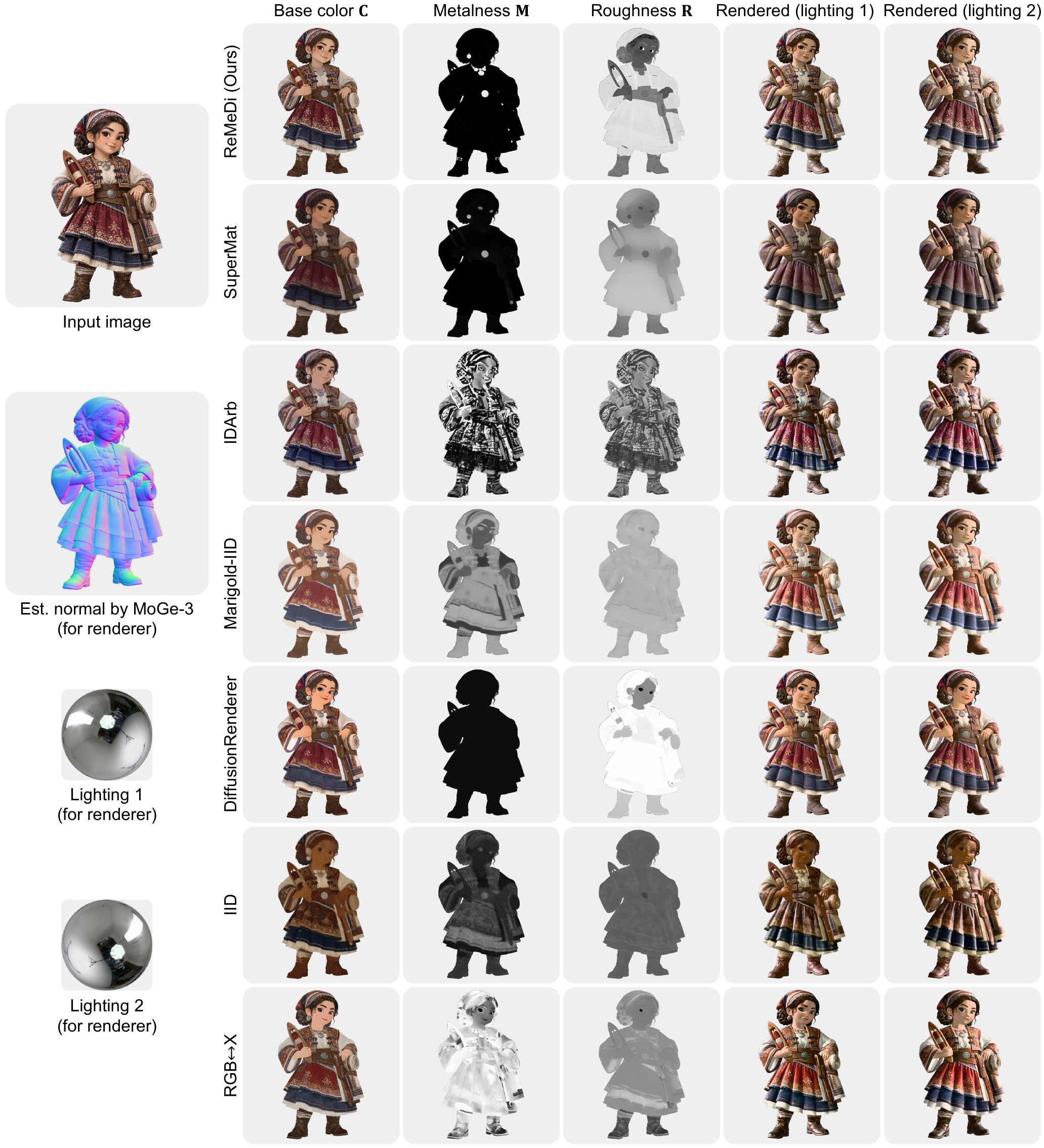}
\caption{
Qualitative comparison on \wildset{}.
We show the base color, metalness, roughness, and relit results using estimated normals.
}
\label{fig: appendix comparison wildtest 3}
\end{figure*}

\begin{figure*}[t!]
\centering
\includegraphics[width=0.99\textwidth]{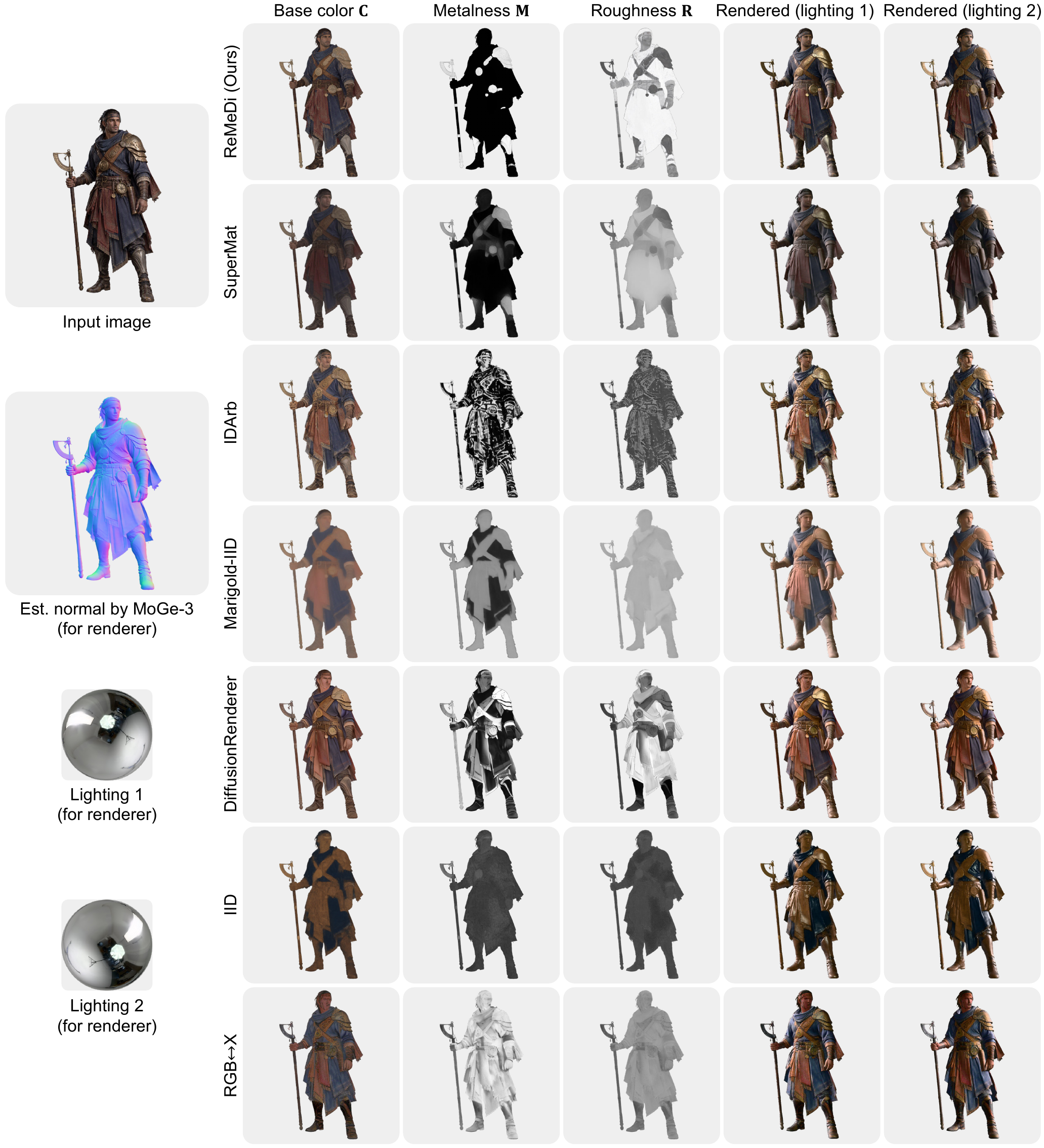}
\caption{
Qualitative comparison on \wildset{}.
We show the base color, metalness, roughness, and relit results using estimated normals.
}
\label{fig: appendix comparison wildtest 4}
\end{figure*}

\begin{figure*}[t!]
\centering
\includegraphics[width=0.99\textwidth]{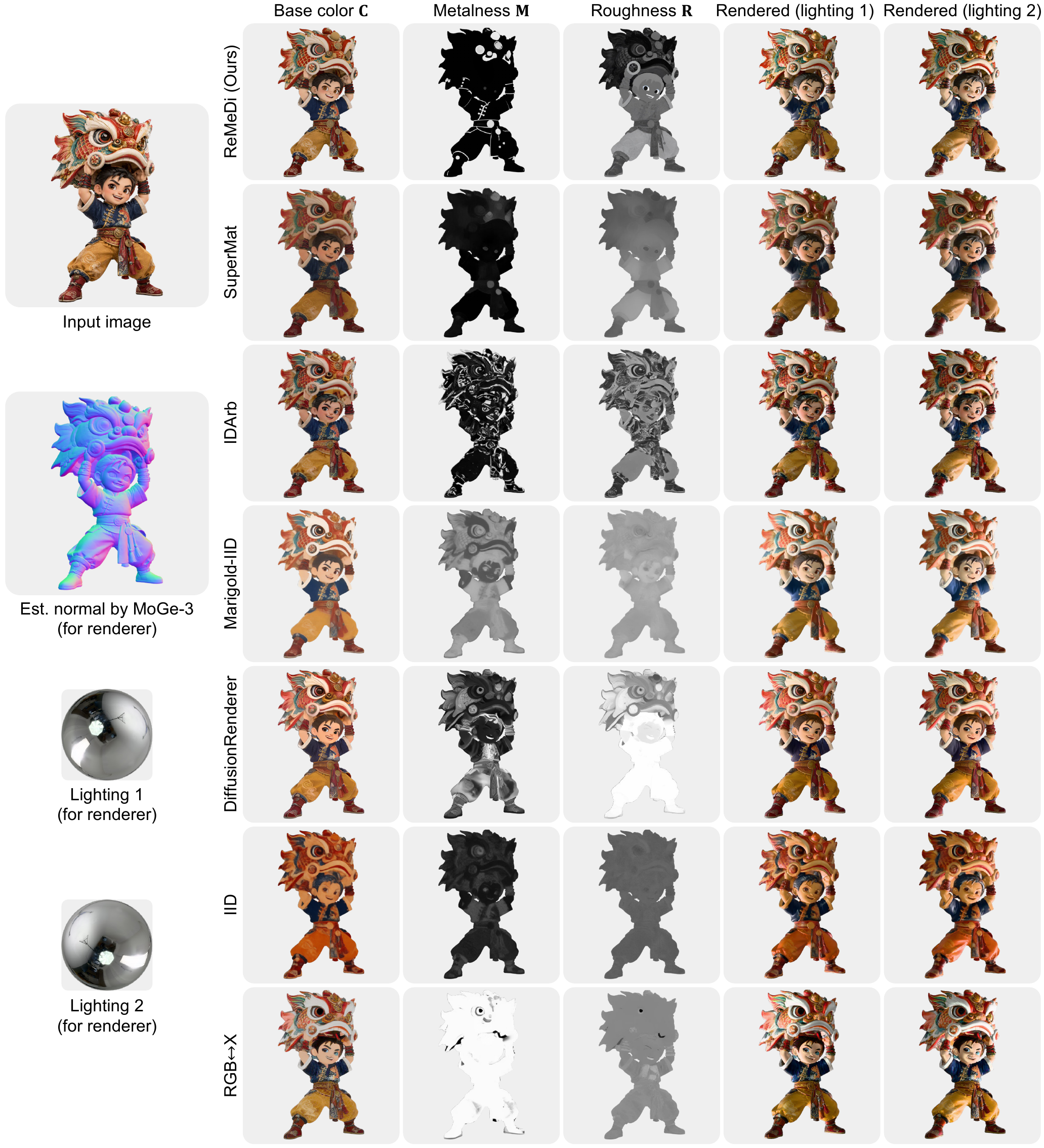}
\caption{
Qualitative comparison on \wildset{}.
We show the base color, metalness, roughness, and relit results using estimated normals.
}
\label{fig: appendix comparison wildtest 5}
\end{figure*}

\begin{figure*}[t!]
\centering
\includegraphics[width=0.99\textwidth]{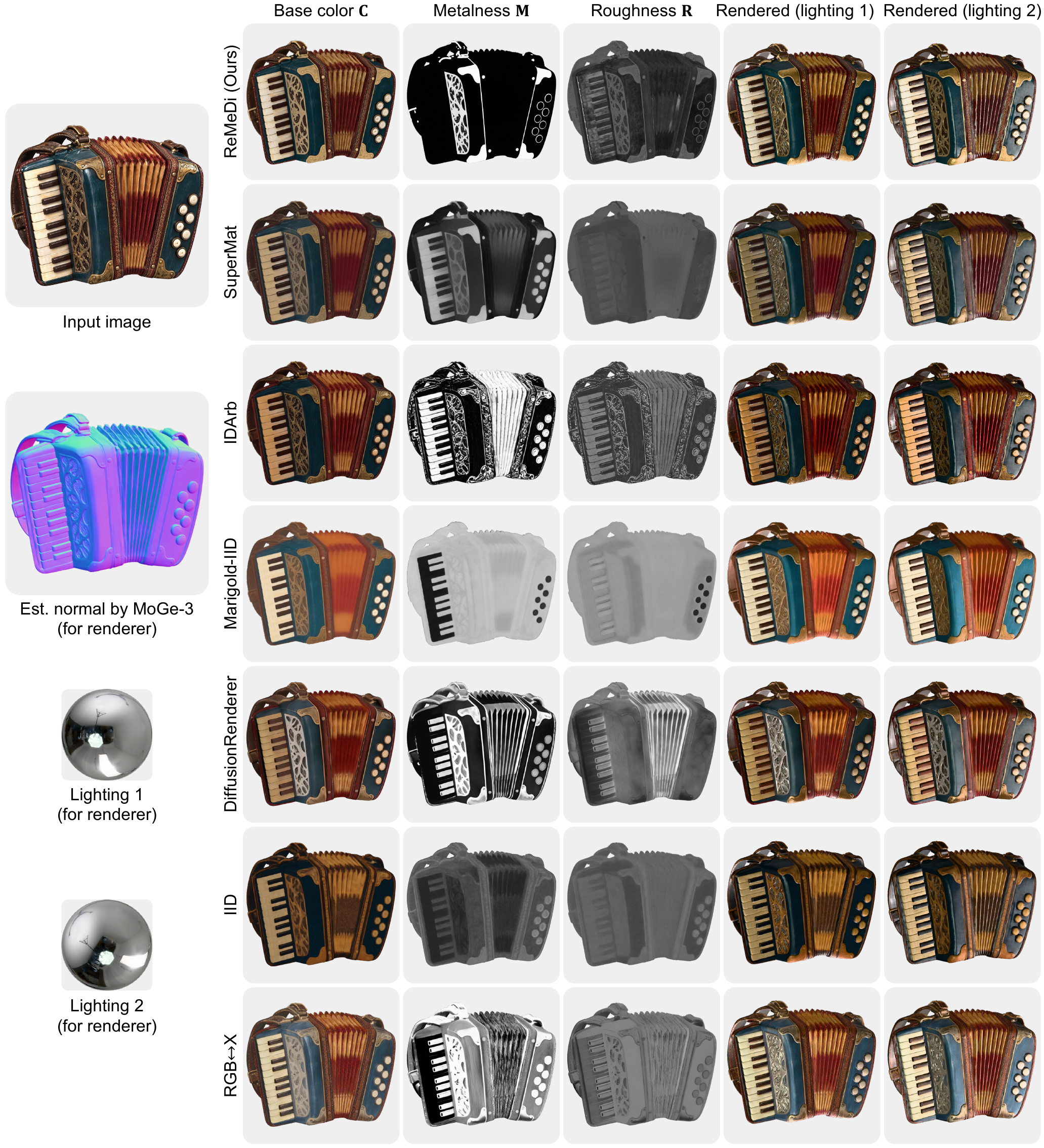}
\caption{
Qualitative comparison on \wildset{}.
We show the base color, metalness, roughness, and relit results using estimated normals.
}
\label{fig: appendix comparison wildtest 7}
\end{figure*}

\cref{fig: appendix comparison syntest 1,fig: appendix comparison syntest 2,fig: appendix comparison syntest 3,fig: appendix comparison wildtest 1,fig: appendix comparison wildtest 2,fig: appendix comparison wildtest 3,fig: appendix comparison wildtest 4,fig: appendix comparison wildtest 5,fig: appendix comparison wildtest 7} show the results of ReMeDi and comparing methods on \synset{} (\cref{fig: appendix comparison syntest 1,fig: appendix comparison syntest 2,fig: appendix comparison syntest 3}) and \wildset{} (\cref{fig: appendix comparison wildtest 1,fig: appendix comparison wildtest 2,fig: appendix comparison wildtest 3,fig: appendix comparison wildtest 4,fig: appendix comparison wildtest 5,fig: appendix comparison wildtest 7}), including estimated materials and their relit results under two novel lighting conditions.
\cref{fig: appendix comparison syntest 1} shows a globe with a metal frame.
\cref{fig: appendix comparison syntest 2} shows an Asahi Pentax SLR camera with a silvery body.
\cref{fig: appendix comparison syntest 3} shows an Iron Man helmet.
\cref{fig: appendix comparison wildtest 1} shows a shield with an antique brass eagle relief.
\cref{fig: appendix comparison wildtest 2,fig: appendix comparison wildtest 3,fig: appendix comparison wildtest 4,fig: appendix comparison wildtest 5} show characters with metal accessories or armors.
\cref{fig: appendix comparison wildtest 7} shows an accordion with metal trims and grilles.
Across these examples, the base color of ReMeDi matches its metalness.
Metal parts, such as the frame of the globe, the body of the camera, the Iron Man helmet, and the trims of the accordion, receive a bright base color that is free of baked-in highlights and serves as specular reflectance, whereas dielectric parts receive a base color that preserves the details of the input and serves as diffuse albedo.
The metalness of ReMeDi is nearly binary and follows part boundaries, and its roughness is smooth within each part, while the metalness and roughness of other methods are often noisy, follow the texture of the input instead of the material, or take intermediate values over whole objects.
As a result, the relit results of ReMeDi keep the look of the input, \eg, the luster of the metal parts, while other methods often make metal parts look dull or overly dark, or give dielectric parts a metallic sheen.

\mypara{Ablation study.}

\begin{figure*}[t!]
\centering
\includegraphics[width=\textwidth]{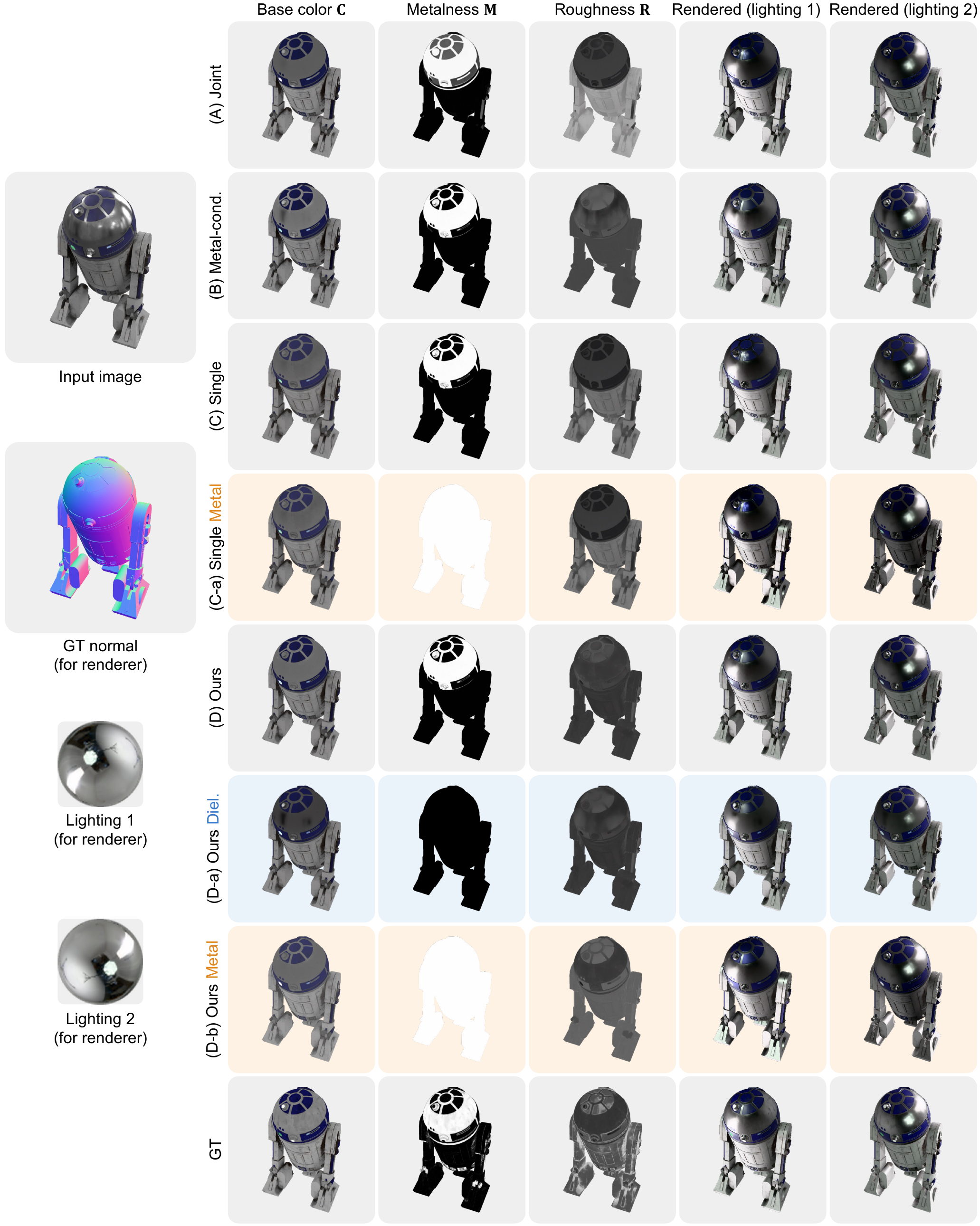}
\caption{
Qualitative ablation results on \synset{} with GT.
For (D) Ours, we show the maps under the metal and the dielectric hypothesis with fixed metalness.
For (C) Single, we show only the maps under the metal hypothesis, as its results under both hypotheses are almost identical.
}
\label{fig: appendix ablation syntest 1}
\end{figure*}

\begin{figure*}[t!]
\centering
\includegraphics[width=\textwidth]{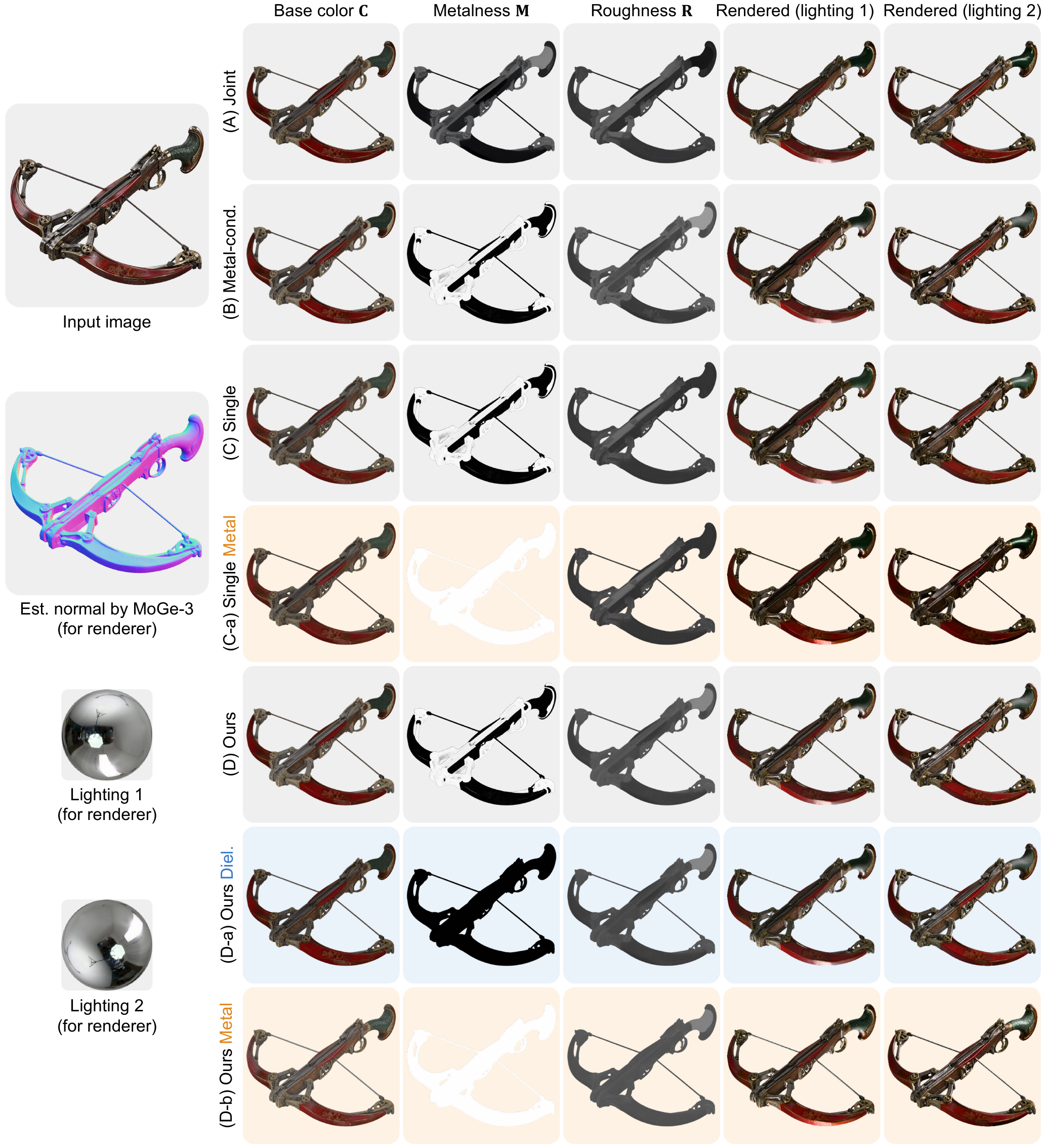}
\caption{
Qualitative ablation results on \wildset{}.
For (D) Ours, we show the maps under the metal and the dielectric hypothesis with fixed metalness.
For (C) Single, we show only the maps under the metal hypothesis, as its results under both hypotheses are almost identical.
}
\label{fig: appendix ablation wildtest 1}
\end{figure*}

\cref{fig: appendix ablation syntest 1,fig: appendix ablation wildtest 1} provide further visual comparisons between ablated variants.
As shown in \cref{fig: appendix ablation syntest 1}, the full model not only estimates base color with more accurate brightness, but also shows stronger ability to remove metallic highlights in the input image, while the (B) metalness-conditioned model and (C) single network both bake the specular shading into the base color.
In both examples, the blended base color maps of the full model match the metalness maps and are free of baked lighting effects.

\end{document}